\documentclass[final,twocolumn]{elsarticle}

\usepackage{lineno}
\usepackage{eqnarray,amsmath}
\usepackage{amssymb}
\usepackage{multirow}
\usepackage{color}
\usepackage{soul} 
\usepackage{bbold}
\usepackage{algorithm2e}
\usepackage{graphicx}
\usepackage{epsfig}
\usepackage{subfig}
\usepackage{multirow}
\usepackage{mathtools}
\usepackage{picins}
\usepackage{hyperref}
    \hypersetup{
        colorlinks   = true,
        citecolor    = blue
    }
\def\v{\checkmark} 

\graphicspath{{./images/}}

\newcommand{\OTHERWISE}{\text{otherwise}} 

\newcounter{Examplecount}[section]

\newcommand{\mat}[1]{\mbox{\boldmath{$#1$}}}

\newcommand{\ssx}{\textbf{s}}

\newcommand{\xx}{\textbf{x}}

\newcommand{\set}[1]{\mathcal{#1}}

\newcommand{\Dcj}{\set{D}}

\newcommand{\Lcj}{\set{L}}

\newcommand{\Xcj}{\set{X}}
\newcommand{\Ycj}{\set{Y}}

\newcommand{\THETA}{\mat{\theta}}

\newcommand{\PARTCTWO}[4]{
\begin{cases} 
		#1, & #2 \\ 
		#3, & #4 
\end{cases}
}

\newcommand{\ARGMIN}[2]{\mathop\text{argmin}_{#1}\left\{#2\right\}} 
\newcommand{\ARGMAXL}{\mathop\text{argmax}}

\makeatletter
\renewcommand*\env@matrix[1][*\c@MaxMatrixCols c]{%
  \hskip -\arraycolsep
  \let\@ifnextchar\new@ifnextchar
  \array{#1}}
\makeatother

\DeclarePairedDelimiter\ceil{\lceil}{\rceil}

\modulolinenumbers[5]

\journal{Neurocomputing}

\begin{document}

\begin{frontmatter}

\title{Action Anticipation for Collaborative Environments: \\
The Impact of Contextual Information and Uncertainty-Based Prediction}

\author[ufes]{Clebeson Canuto}
\ead{clebeson.canuto@gmail.com}

\author[ist]{Plinio Moreno}
\ead{plinio@isr.tecnico.ulisboa.pt}

\author[ufes]{Jorge Samatelo}
\ead{jorge.samatelo@ufes.br}

\author[ufes]{Raquel Vassallo}
\ead{raquel@ele.ufes.br}

\author[ist]{Jos\'e Santos-Victor}
\ead{jasv@isr.tecnico.ulisboa.pt }

\address[ufes]{Room 20, CT-II, Department of Electrical Engineering, Federal University Esp\'irito Santo, Av. Fernando Ferrari, 514, Goiabeiras, 29075-910, Vit\'oria - ES - Brazil}
 
\address[ist]{ Floor 7, North Tower, Institute for Systems and Robotics, Instituto Superior T\'ecnico, Universidade de Lisboa, Av. Rovisco Pais, 1, 1049-001, Lisbon, Portugal}

\begin{abstract}
To interact with humans in collaborative environments, machines need to be able to predict (i.e., anticipate) future events, and execute actions in a timely manner. However, the observation of the human limb movements may not be sufficient to anticipate their actions unambiguously. In this work, we consider two additional sources of information (i.e., context) over time, gaze, movement and object information, and study how these additional contextual cues improve the action anticipation performance. 
We address action anticipation as a classification task, where the model takes the available information as the input and predicts the most likely action. We propose to use the uncertainty about each prediction as an online decision-making criterion for action anticipation. Uncertainty is modeled as a stochastic process applied to a time-based neural network architecture, which improves the conventional class-likelihood (i.e., deterministic) criterion. 
The main contributions of this paper are four-fold: 
(\textit{i}) We propose a novel and effective decision-making criterion that can be used to anticipate actions even in situations of high ambiguity;
(\textit{ii}) we propose a deep architecture that outperforms previous results in the action anticipation task when using the Acticipate collaborative dataset; 
(\textit{iii}) we show that contextual information is important to disambiguate the interpretation of similar actions; and 
(\textit{iv}) we also provide a formal description of three existing performance metrics that can be easily used to evaluate action anticipation models.
Our results on the Acticipate dataset showed the importance of contextual information and the uncertainty criterion for action anticipation. We achieve an average accuracy of $98.75\%$ in the anticipation task using only an average of $25\%$ of observations. Also, considering that a good anticipation model should perform well in the action recognition task, we achieve an average accuracy of $100\%$ in action recognition on the Acticipate dataset, when the entire observation set is used.
\end{abstract}


\end{frontmatter}

\section{Introduction}
Humans have the natural ability to interact with each other and perform joint tasks. Part of this ability is due to their capacity of perceiving the environment and recognizing patterns that help them anticipate the actions of others, and thus make better decisions. Similarly, artificial machines need this capacity of anticipating actions, to act accordingly and achieve an effective interaction with humans~\citep{duarte2018action}.  
 
Action anticipation and action recognition are two different tasks. The ``action recognition'' task is based on a model that uses an entire sequence of information, which represents one performed action, to associate the observed action to one possible action class \citep{action_rec_survey2018human}. If the decision-making depends on the entire action, it can only be performed after the action is completely executed. However, this approach is not suitable for systems that manage risks or perform joint tasks with humans. For instance, in a situation where a self-driving car approaches a pedestrian, it must perceive whether the pedestrian will cross the road in time, in order to safely stop or deviate the car if necessary. In this scenario, the model must not only recognize actions but, more importantly, must anticipate them~\citep{liu2020pedestrian}. 

Action anticipation consists of classifying an action even before it occurs, by using the partial information provided up to a certain moment in time. Usually, an anticipation model is more complex than a recognition one. This comes from its capacity to classify actions based on an incomplete sequence of data, which makes the choice of the correct class more uncertain. Ideally, every anticipation model should be capable of recognizing actions; on the other hand, not every recognition model would be able to anticipate them. 

 In the last few years, deep learning has achieved the state-of-the-art results in many tasks, such as image recognition~\citep{resnet2016deep,krizhevsky2012alexnet,simonyan2014vgg16}, natural language processing~\citep{devlin2019bert,vaswani2017attention} and action/activity recognition~\citep{karpathy2014video,simonyan2014twostrem,carreira2017i3d}. Some works, like~\citep{bilen2017action_movement,rodriguez2018dynamic_image,choutas2018potion}, represent an action by estimating the movement of the involved actors (i.e., users). In the case of simple and unambiguous actions, the movement can be sufficient for a successful recognition/anticipation task. However, in the case of more complex and ambiguous actions, it would not be enough to recognize/anticipate successfully, mainly when the information about objects, persons, environment configuration, movements performed previously, are important for recognizing or anticipating actions. Furthermore, some details during action anticipation, such as objects' position, the relation between hands and object/person and the type of object manipulated, can offer as much or even more information than only movement. As such, using only movement, the model rules out the context, a critical information that can help characterize the actions.
 
 Regarding the action recognition task, the two-stream approaches \citep{simonyan2014twostrem,carreira2017i3d,kwon2018two_stream} are the most successful, because they use movement as the main source of information to describe each action, and they use the context as additional information that can help characterize each class individually. In these solutions, the movement is the optical flow calculated between sequential images, and the contextual information \cite{baradel2018object} is extracted implicitly by CNNs (Convolutional Neural Networks)\citep{lecun1995cnn}. However, to obtain the implicit contextual information from images in a self-supervised manner, the training procedure of the CNN models requires large datasets to achieve good results. As a consequence, the two-stream approaches are not effective when solving problems provided by small datasets, such as those commonly used for human-human or human-robot collaboration.  
 
 Analyzing from another perspective, even achieving satisfactory results in their experiments, the aforementioned works are not crystal clear about how one could use their solutions in a real-time situation, once they measure the model performance using accuracy or observation ratio. They do not discuss how to handle action anticipation or what kind of function must be used as the decision-making criterion. Due to the absence of such discussion, it is unclear how to use this approach in a real application, where the data is continuously generated, as in a video streaming.
 
Another problem of most deep learning solutions is their overconfidence in their predictions. A deterministic model will always provide a prediction, even when there is a high uncertainty about the correct class, and the final decision becomes unclear. A trustworthy model should assess its uncertainty about each prediction and provide the system with the possibility of making more reliable decisions. 
 
In this work, we focus on context-based action anticipation, but with small datasets. Thus, instead of implicitly learning the visual context, we define the contextual information in an action anticipation problem. With this in mind, we used the Acticipate dataset \citep{duarte2018action}, where one person hands an object over to another one, and receives it back. For this dataset, the dyadic interaction task requires the future prediction (i.e., anticipation) of the arm and head motion, gaze and object position. A previous work \cite{schydlo2018anticipation} has shown that, using the eye gaze and the 3D pose of the main character in the Acticipate dataset, a time-based deep learning architecture is able to anticipate his actions. As defined in \citep{Dey1999}, context is any information that can be used to characterize an entity. Therefore, when considering the 3D pose/movement as the entity that represents an action, the eye gaze in \citep{schydlo2018anticipation} can be seen as context information. 

Now, to further investigate the importance of context in the task of anticipating actions, we have increased the complexity in the Acticipate dataset, extending its number of actions. To do that, we divided previous actions to create the new actions \textit{receive} and \textit{pick}, which add ambiguity into the actions \textit{give} and \textit{place}, correspondingly. We also consider an additional element of context information, the position of the handled object. Finally, instead of using 3D pose and gaze as in~\citep{schydlo2018anticipation}, we use only information taken from RGB images. Such restriction makes our proposal more general and less dependent on intrusive and/or expensive sensors.

Also, to investigate the possibility of using the uncertainty to provide a more reliable decision, we propose a context-aware model based on a recurrent neural network with an adaptive threshold. This threshold is calculated via an uncertainty metric and represents a decision-making criterion for action anticipation. The use of uncertainty significantly contributes to attenuate the overconfidence problem often faced by models trained with small datasets. 

In summary, the main contributions of this paper are the following:
\begin{itemize}
\item 
We propose a novel and effective decision-making criterion that can be used to anticipate actions even in situations of high ambiguity. The proposed approach aims to minimize the model's uncertainty instead of maximizing its class probabilities. Therefore, by applying a proper threshold over the uncertainty, the decision about whether an action should be anticipated or not can be done.
\item
We show the importance of context information to disambiguate similar actions.
\item 
We propose a deep architecture that uses less information than \citep{schydlo2018anticipation}, and outperforms the results in action anticipation task using Acticipate dataset. This result holds even in the case of its extended number of actions, which are more ambiguous than the original ones \citep{duarte2018action}.
\item 
We also provide a formal description of three existing performance metrics that can be easily used to evaluate action anticipation models.

\end{itemize}

To build a better understanding about our proposal, the next sections will cover, respectively: 
the related works (Sec.~\ref{sec:related_works}); action anticipation background and related problems (Sec.~\ref{sec:action_anticipation}); 
the methodology of this work, including the hypotheses raised and its main contributions (Sec.~\ref{sec:intuitions}); Bayesian neural networks and uncertainty (Sec. \ref{sec:Bayesian_networks});
our proposed approach (Sec. \ref{sec:proposal}); 
experiments (Sec.~\ref{sec:experiments}~and~\ref{sec:results}, for results and discussions); and finally, conclusions and future works (Sec.~\ref{sec:conclusions}).

\section{Related Works}
\label{sec:related_works}

In the last few years, action anticipation has been addressed by many researchers \citep{wang2019ant,pirri2019ant,agethen2019ant,shi_rbf_2018actant,hu2018ant_early,sadegh2017ant_lstm} due to its importance to perform an effective interaction.

In \citep{shi_rbf_2018actant}, the authors proposed to decrease the dimensionality on RNNs by allowing the sharing of weights, and improve the temporal representation of an action by using an RBF kernel (Radial Base Function) over the hidden-state of an LSTM network. 
They proposed to feed an LSTM with features extracted by a CNN. Next, they applied an RBF over the LSTM hidden states, and lastly, the RBF outcome is given as input to a Multilayer Perceptron (MLP). The authors use between $20\%$ and $50\%$ of a video to predict the next features and then perform the anticipation. 

In \citep{rodriguez2018dynamic_image}, the authors use a convolutional auto-encoder network to predict the next movement of a video. Such movement is generated by a ranking loss function, applied over the difference between consecutive images in a sequence, and is stored in a still RGB image called Dynamic Image~\citep{bilen2016dynamic_image}. With a Markov assumption, after generating a sequence of dynamic images using $S$ frames for each one, the model generates the next $k$ dynamic images, where $k \geq 1$. Further, those images feed a model that outputs the probability distribution over action classes. A drawback of the two previous works is to use movement as the only source of information to represent an action, which can harm the prediction of actions that are related not only to movement but also to context information.

In \citep{basura2017encouraging} is proposed a model to anticipate actions based only on RGB images. The authors use as feature extractor the pre-trained CNN VGG16 and, as the classifier, two LSTMs that predict the classes corresponding to each input frame of a video. A similar approach is also presented in \cite{basura2016deep}.

LSTM is also used to anticipate actions of car drivers by using only RGB images \citep{gite2019early} or in combination with GPS information \citep{gite2019earlygps}. Other approaches, as \citep{wang2019early_gan}, use Generative Adversarial Networks (GAN) to predict future images and then anticipate the action, or more sophisticated architectures, as in \citep{liu2020pedestrian}, that uses Convolutional Graphical Models (CGM) to predict when a pedestrian will cross the road.

Despite these works present good results in terms of accuracy at each observation time, none of them explains how action anticipation should be performed in a real scenario, when it is not possible to know the size of the input sequence. They did not discuss what kind of decision-making criteria could be used in such a situation.

Even in works as \citep{liu2019onlineskeleton}, which aim to anticipate action in online videos, the authors only reported the accuracy at each observation, but nothing about how to make decisions. Only a couple of works address this question. For instance, in \citep{jain2015car,jain2016car}, the authors use a threshold over the probability distribution provided by an HMM (Hidden Markov Model) to anticipate maneuvers of drivers. However, as discussed in \citep{jain2015car}, this approach faces problems in ambiguous situations, where it is not possible to be sure about the action to be anticipated, even when the probability exceeds the specified threshold.

Many of the approaches mentioned above are not suitable for small datasets, since the high capacity of their models can lead to overfitting. Therefore, \citep{schydlo2018anticipation} proposes a different method to anticipate action in the Acticipate dataset - a small collaborative dataset used to understand the role of gaze on action anticipation \citep{duarte2018action}, as discussed in Sec.~\ref{sec:intuitions}. Their approach consists of feeding an LSTM cell with a 3D pose (Motion Capture-MoCap information) and gaze (fixation points), and then pass the LSTM output through a softmax classifier. They trained two models with different observations: one with only 3D pose and another with 3D pose plus eye gaze. When the model uses the pose and gaze information, the authors concluded that the actions in the dataset could be anticipated 92ms before. This result showed the importance of using not only movement information (here, the evolution of the pose in time) to anticipate actions. 
However, the authors did not notice that their model did not recognize all the actions ($100\%$ of action recognition accuracy) even after seeing the whole sequence. Their results for action anticipation were shown based only on one action sample. More conclusive results should present statistics for all classes in the entire dataset. In complement, they also did not provide an answer to when a model must anticipate an action. From their comments, we presume that it may be done using a threshold on the probability value, as mentioned in \citep{jain2015car, jain2016car}.

After these explanations, our main objectives in this work are:
  \begin{itemize}
      \item propose a model that improves results in \citep{schydlo2018anticipation} even when using only RGB images;
       \item present how context can be used in a neural network architecture to improve action anticipation;
       \item present in detail how to anticipate an action using a threshold value as a decision-making criterion; and 
       \item propose the use of uncertainty as an effective threshold value that improves action anticipation.
  \end{itemize}

\section{Action Anticipation Background}
\label{sec:action_anticipation}

In this section, we describe the definition adopted here for action anticipation, its main properties, and how we address the problem. 
We can divide the works that try to solve the anticipation task into two main groups: 
(\textit{i}) early action prediction, where an action must be predicted before it is fully executed \citep{bloom2017lineareraly,hu2018ant_early,wang2019early,wang2019early2,ji2018oneearly}; and
(\textit{ii}) event anticipation, where an event must be predicted before it starts \citep{felsen2017event_pred,neumann2019event_pred,neumann2019event_pred}.  In this work, ``action anticipation'' is understood as in the first set of works: early action prediction by using sequential features.

\subsection{Problem Definition}

First of all, it is essential to formally define the action anticipation task. Let 
$\Xcj = \{\xx_1,\xx_2,\cdots, \xx_{N} \mid \xx_t \in \mathbb{R}^{d \times 1}\}$ be a sequence with $N$ observations that represents the execution of a specific action $y \in \Ycj$, where $\Ycj$ is a set with $d$ action classes. Here, $\xx_t$ represents an observation taken at time $t$. Now, considering that $\Xcj_{t_1:t_2}$ represents an indexed sequence composed by the observations taken between time $t_1$ and $t_2$, we define a model $M$ for action classification problem as a mapping function parametrized by $\THETA$ that receives as input $\Xcj_{1:t}$($t$ observations from $\Xcj$) and return as output the vector of probability scores $\ssx \in [0,1]^{d \times 1}$, representing the probability that sequence $\Xcj$ belongs to each action class.

\begin{equation}
\label{eq:rec_model}
\ssx = M(\Xcj_{1:t},\THETA).
\end{equation}

In action recognition tasks, the model $M$ has all the observations of the sequence $\Xcj$ ($t =N$) available to generate the probability score $\hat{\ssx}$. On the other hand, for an action anticipation task, the action is not completely executed, thus only an initial part of $\Xcj$ is available ($t<N$) so that $M$ can infer $\hat{\ssx}$.

In Eq.~\eqref{eq:rec_model} the parameter $\THETA$ can be found by solving the following optimization problem:
\begin{equation}
\label{eq:rec_optimization}
\hat{\THETA} = \ARGMIN{\THETA}{\Lcj(\THETA,\Dcj)}
\end{equation} 
\noindent where $\Dcj = \{(\Xcj^{(1)},y^{(1)}),(\Xcj^{(2)},y^{(2)}),\cdots,(\Xcj^{(k)},y^{(K)})\}$ is the training set, with each pair $(\Xcj^{(i)},y^{(i)})$ representing an action sequence and its respective label, $K$ is the number of sequences in the training set, and $\Lcj$ is a loss function. 

During the prediction time, we do not know the value of $N$, and thus we do not know when the action will end. Therefore, at each time $t$, $M$ only uses the observed current sequence, $\Xcj_{1:t}$, and a function $g$ is in charge of predicting the action class at instant $t$.
\begin{eqnarray}
\label{eq:prediction}
\hat{\ssx} &=& M(\Xcj_{1:t},\hat{\THETA})\nonumber\\
\hat{y} &=& g(\hat{\ssx}). 
\end{eqnarray}

For action recognition tasks, the discriminant function $g$ can be defined as:
\begin{equation}
\label{eq:thres_max_prob}
g(\hat{\ssx}) = \ARGMAXL(\hat{\ssx} ), 
\end{equation} 
because the model $M$ is more confident about the probability score assigned to $\hat{\ssx}$. On the other hand, for action anticipation tasks, since $M$ uses only part of the observations, when the distribution $\hat{\ssx}$ is close to a uniform distribution, one can not be certain about the correct class. Hence, Eq.~\eqref{eq:thres_max_prob} is not an adequate discriminant function to anticipate actions. 

In this way, a better option is to use a discriminant function with a threshold parameter $p$, as presented in Eq.~\eqref{eq:anticipation_threshold}. 
\begin{equation}
\label{eq:anticipation_threshold}
g(\hat{\ssx} ,p) = \PARTCTWO{\ARGMAXL(\hat{\ssx})}{h(\hat{\ssx} ) > p}{-1}{\OTHERWISE}.
\end{equation}

Once $p$ is specified as a probability value, $h$ can be defined as:
\begin{equation}
\label{eq:prob_thres}
h(\hat{\ssx} ) = max(\hat{\ssx} )
\end{equation}

In Eq.~\eqref{eq:anticipation_threshold}, a value of $p \geq 0.9$ means that the model is highly certain about its prediction, and the action can be anticipated, which favors the use of such a model in real-time. On the other hand, when it returns $-1$ means that it is not certain about the correct class and needs more observation to improve its certainty.

\subsection{Evaluation metrics}
After determining how to anticipate an action, it is essential to decide how to ascertain the quality of the model $M$. Therefore, we formally describe three existing metrics that can be used in anticipation benchmark experiments: (\textit{i}) accuracy at each observation ratio, (\textit{ii}) anticipation accuracy and (\textit{iii}) expected observation ratio.

\noindent\textbf{Accuracy at each observation ratio.} Considering that each sequence $\Xcj$ can have a different length $N$, this metric helps evaluate all sequences in a normalized time scale. Thereby, the success ratio when anticipating an action after a observation ratio $r$, with an anticipation threshold $p$, can be calculated as follows:
\begin{equation}
\label{eq:acc_duration}
ACC(r) = \frac{1}{K}\sum^{K}_{i=1} pred(\Xcj^{(i)}_{1:\ceil{r\times N}},y^{(i)},p), 
\end{equation}

\noindent where,
\begin{equation}
\label{eq:acc_duration2}
pred(\Xcj_{1:t},y,p) = \PARTCTWO{1}{g(M(\Xcj_{1:t}),p) = y}{0}{\OTHERWISE}.
\end{equation}

In Eq.~\eqref{eq:acc_duration}, in terms of $r$, $t= \ceil{r\times N} \ \forall \ r \in (0,1]$, where $N$ is the number of observations in a sequence $\Xcj^{(i)}$. However, in terms of $t$, $r=t/N \ \forall \ t \in \{{1,2,...,N}\}$.
 
\noindent\textbf{Anticipation accuracy.}
In a real-time situation, the model can not access the label of each observation. So, the evaluation of the anticipation model during training must be performed when the model makes its first prediction for each sequence. In this sense, this classification metric measures the success ratio of the model $M$ when anticipating actions by the first time. It is calculated as the average accuracy of each classification. Therefore, when using this metric, we do not regard in which observation the action was predicted but whether it was predicted correctly. Eq. \ref{eq:acc_ant_act} presents how it is calculated,

\begin{equation}
\label{eq:acc_ant_act}
ACC_{act} = \frac{1}{K}\sum^{K}_{i=1}\sum^{N-1}_{t=1} I(t) pred(\Xcj^{(i)}_{1:t},y^{(i)},p), 
\end{equation}

\noindent where,

\begin{equation}
\small
\label{eq:iloc2}
I(t) = \begin{cases}
                0, & ((t = 1) \wedge (g(M(\Xcj_{1:t}),p) = -1)) \\ & \vee (I(t-1) = 1)\\
                1, & \OTHERWISE
          \end{cases}
\end{equation}

\noindent where $I$ is an indicator function that disable predictions based on whether the anticipation has already occurred or not.

\begin{figure}[]
    \centering
    \includegraphics[width = 1.\linewidth]{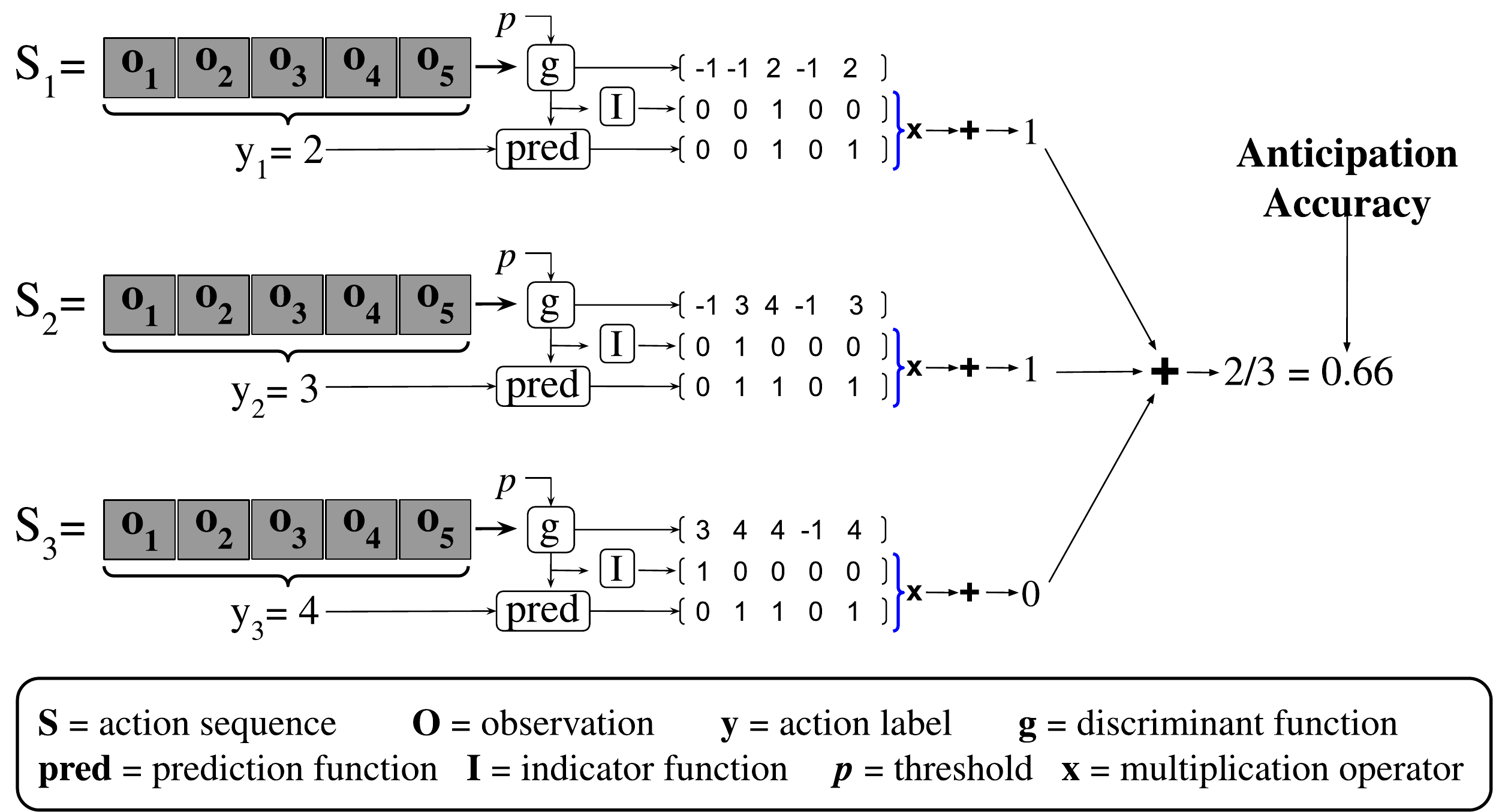}
    \caption{Graphical example of how to calculate the anticipation accuracy using Eq.~\eqref{eq:acc_ant_act}.}
    \label{fig:acc_ant_act}
\end{figure}

A detailed example of how to apply this metric is given in Fig.~\ref{fig:acc_ant_act}.

\noindent\textbf{Expected observation ratio.}
This measurement focuses on the expected amount of observations necessary to anticipate an action correctly. It can be implemented according to Eq.~\eqref{eq:expected_duration}. Note that when the model is correct, it receives the value $t$, which corresponds to the observation where the prediction is executed. However, when it misses the anticipation, it is penalized by receiving the sequence size $N$. 
\begin{equation}
\label{eq:expected_duration}
E_{obs}\ = \frac{1}{K}\sum^{K}_{i=1} obs(\Xcj^{(i)},y^{(i)},p),
\end{equation}

where,
\begin{eqnarray*}
obs(\Xcj,y, v) &=& \frac{1}{N}min(\{f_{pred}(\Xcj_{1:t},y,p,t,N)\}_{t=1}^{N})\\
f_{pred}(\Xcj_{1:t},y,p,t,N) &=& \PARTCTWO{t}{pred(\Xcj_{1:t},y,p)=1}{N}{\OTHERWISE}.
\end{eqnarray*}

\section{Methodology}
\label{sec:intuitions}
This work aims to show the influence of context in the anticipation task and to use uncertainty as a decision-making criterion in a collaborative environment. To do this, we use a  controlled dataset that contains, by each frame, the action performed and the corresponding context information. Therefore, to understand how our intuitions have arisen and resulted in our proposal, it is necessary to analyze the used dataset and thus realize how the questions came up.

\subsection{Acticipate dataset}

\begin{figure} [h!]
\centering
\subfloat[place right]{\includegraphics[width =0.3\linewidth]{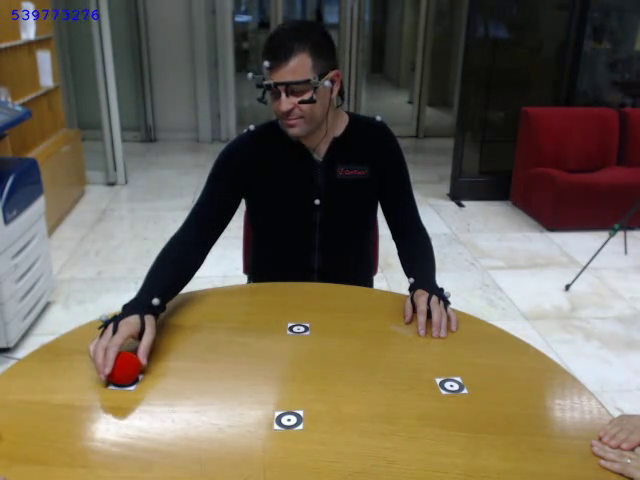}} \,
 \subfloat[place middle]{\includegraphics[width =0.3\linewidth]{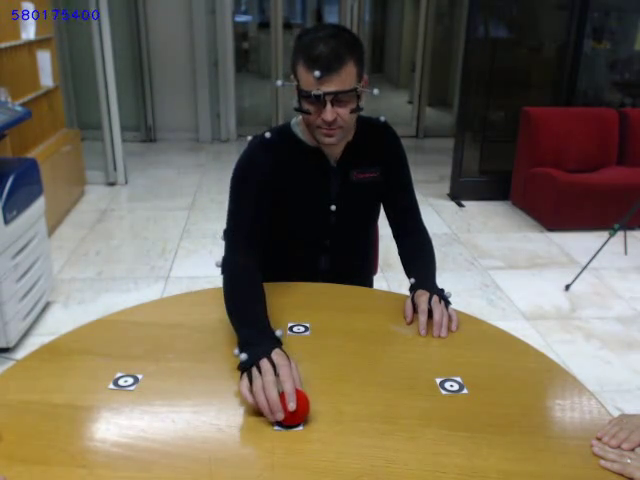}}\,
 \subfloat[place left]{\includegraphics[width =0.3\linewidth]{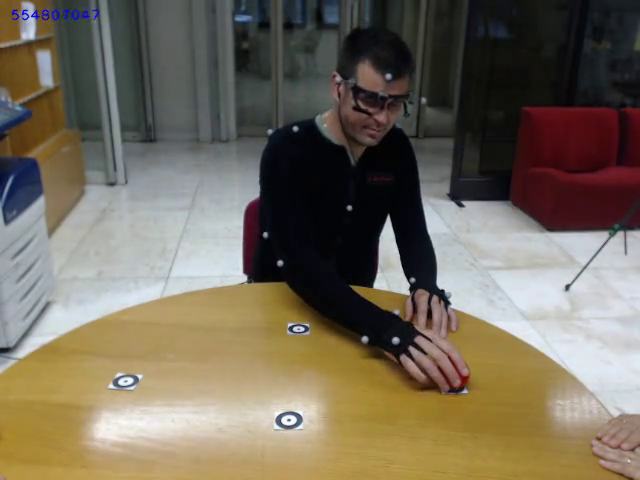}}\,
 
 \subfloat[give right]{\includegraphics[width =0.3\linewidth]{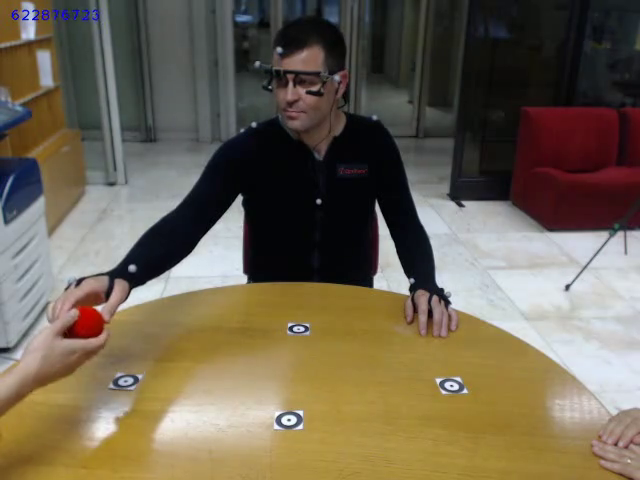}}\,
 \subfloat[give middle]{\includegraphics[width =0.3\linewidth]{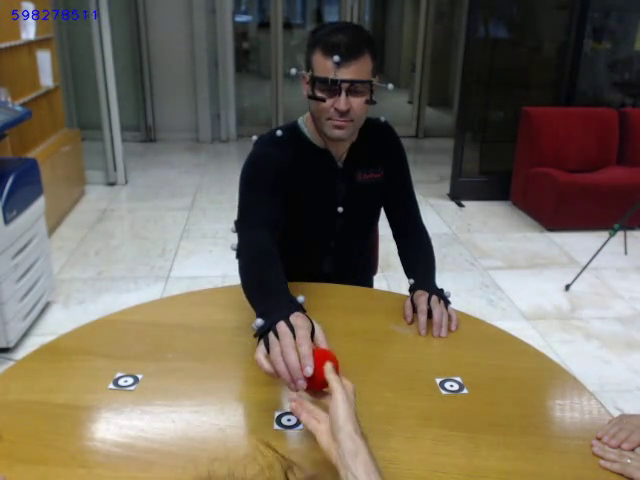}}\,
 \subfloat[give left]{\includegraphics[width =0.3\linewidth]{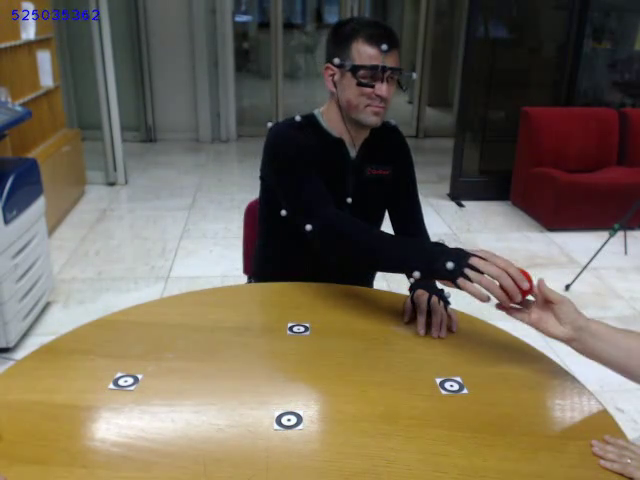}}\,
 
 \subfloat[object starting point]{\includegraphics[width =0.3\linewidth]{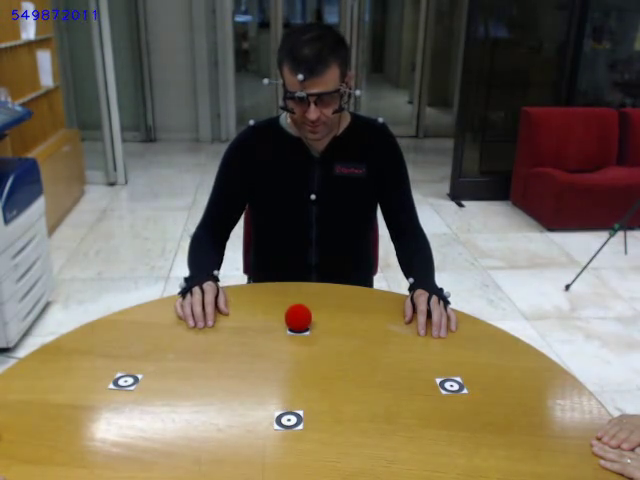}}

\caption{Sample of each action in Acticipate Dataset and the object starting point}
\label{fig:sample_all_actions}
\end{figure}

The chosen dataset is the Acticipate\footnote{Download: \url{http://vislab.isr.ist.utl.pt/datasets/}}, which was acquired to study the influence of gaze in action and/or intention anticipation~\citep{duarte2018action,schydlo2018anticipation} in a collaborative environment. It comprises $120$ trials, distributed into 6 classes. During the acquisition, the actor was wearing an eye gaze tracker binocular glasses (Pupil Labs eye-tracker \citep{kassner2014pupil}) and a suit with 25 markers. He should perform 6 different actions: \textit{give} an object (left, middle, or right) and \textit{place} an object (right, middle, or left). In the \textit{give} actions, he should give an object (in this case, a small red ball) to one of the three volunteers located on: his right side, left side, or in front of him (middle). In the \textit{place} actions, he should place the same object in one of the three points on the table located at his right, middle (in front of him), or his left. Each action starts with the object placed in a point near to the actor and finishes when the object returns to the same point. As showed by ~\citep{duarte2018action} the gaze is an essential source of information when one wants to anticipate actions. Besides, as we will see in the next section, the object plays a fundamental role when the action becomes more ambiguous. So it is possible to know what kind of context information must be taken into account for each action class during the anticipation process. Fig.~\ref{fig:sample_all_actions} presents a sample of each action and the object starting point.

Each trial consists of 3-dimensional data corresponding to the positions of the markers on the actor's suit, captured by an OptiTrack\footnote{https://optitrack.com/} MoCap system, at 120Hz; 2D gaze fixation point captured by the eye tracker glasses at 60Hz; and an RGB video captured by a camera facing the actor, at 30Hz. The dataset is unbalanced, because every class has a different number of samples: 17 (place right), 23 (place middle), 20 (place left), 24 (give right), 19 (give middle) and 17 (give left). 

In this work, when referring to the Acticipate dataset, we call movement the change of position of both arms.

\subsection{Dataset analyzis}

By analyzing the dataset, it is possible to notice that, in many cases, the movement does not have enough information about action to provide good anticipation. For instance, each action of \textit{place} and \textit{give} has similar movements depending on its direction (left, middle or right). However, after taking into account gaze information, one can notice that the action can be anticipated long before. Gaze indicates whether the user will place the object on the table or give it to a volunteer. As discussed before, if we take the movement as the principal entity of each action, we can consider gaze as a context information (additional information that helps characterize the entity). Now, gaze and movement provide enough data to anticipate actions in this dataset. However, if the dataset was divided into more actions, would the gaze be a sufficient source of information to anticipate them?

\begin{figure} [h!]
\centering
\subfloat[Place middle]{\includegraphics[width =1.0\linewidth]{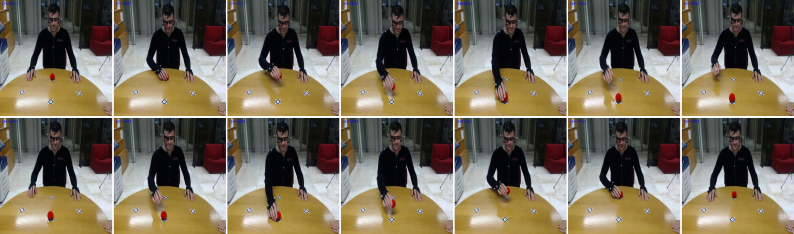}}\\
 \subfloat[Give right]{\includegraphics[width =1.0\linewidth]{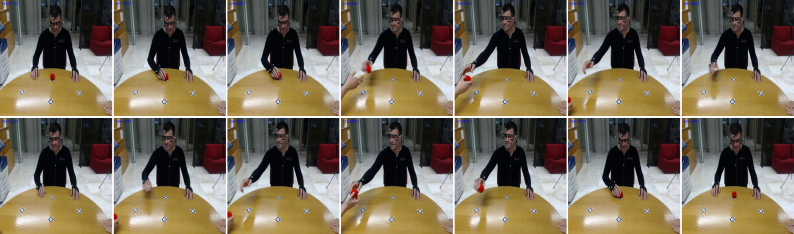}}
\caption{Sample of two actions (place middle and give right) from Acticipate Dataset}
\label{fig:sample_place_give}
\end{figure}

An interesting but not considered characteristic of this dataset is that, once the interaction involves only one object, the actor must place it at its starting point after performing each action. Thus, when he places the object somewhere on the table, he must \textit{pick} it up, and when he gives the object to someone, he must \textit{receive} it back. A simple example of this behavior can be seen in Fig.~\ref{fig:sample_place_give}. In that way, we can extend the dataset from 6 actions to 12 actions: \textit{give}, \textit{place}, \textit{pick} and \textit{receive} (each one with the directions left, right and middle).

\begin{figure} [h!] 
\centering
\subfloat[any action is possible]{\includegraphics[width =0.22\linewidth]{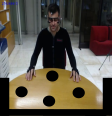}}\,
 \subfloat[any action is possible]{\includegraphics[width =0.22\linewidth]{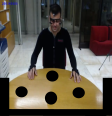}}\,
 \subfloat[place or pick left]{\includegraphics[width = 0.22\linewidth]{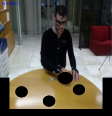}} \,
 \subfloat[place or pick left]{\includegraphics[width = 0.22\linewidth]{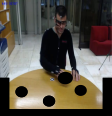}}\,

\subfloat[give or place (any direction)]{\includegraphics[width = 0.22\linewidth]{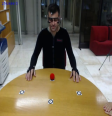}}\,
 \subfloat[pick middle]{\includegraphics[width = 0.22\linewidth]{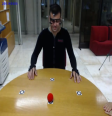}}\,
 \subfloat[place left]{\includegraphics[width = 0.22\linewidth]{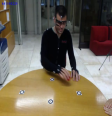}} \,
 \subfloat[pick left]{\includegraphics[width = 0.22\linewidth]{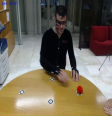}}\,

\caption{Situations in the extended dataset with great ambiguities when it is analyzed only gaze and movement. Any action is possible in (a) and (b). It is necessary to wait for the movement to infer the direction but, even after knowing the direction, it is necessary to observe almost the complete action to distinguish between the actions \textit{give/receive} and \textit{place/pick}. In (c) and (d), the movement starts toward the left side, simultaneously, the gaze is directed to the table. Therefore, the possible action is a \textit{place} or \textit{pick} toward the left direction. For the actions shown in (e)-(h), because the object position is taken into account, the ambiguities can be reduced or even eliminated. In (e) and (f), the number of possible actions is reduced after knowing the object position. In (e), \textit{pick} and \textit{receive} actions are not possible. On the other hand, in (f), only the action \textit{pick middle} is possible. The same occurs in (g) and (h). In (g), the most likely action is \textit{place left}. Finally, in (h), the only possibility is \textit{pick left}. Notice that in (f) and (h), the action is anticipated after observing only one image.}
\label{fig:sample_ambuiguities_object}
\end{figure}

Considering now the extended dataset, if we constrain the analyzes to the movement and gaze (Fig.~\ref{fig:sample_ambuiguities_object} (a-d)), we notice that, even with gaze, it is not possible to perform right anticipation between actions \textit{give}/\textit{receive} or \textit{place}/\textit{pick} when they are toward the same direction. In this case, it is necessary to wait for more observations. 

On the other hand, when applying no constraint on what we can analyze in each image (Fig.~\ref{fig:sample_ambuiguities_object} (e-h)), we can anticipate actions of the extended dataset as fast as in its original configuration. In some cases, as in Fig.~\ref{fig:sample_ambuiguities_object} (f), the action can be anticipated after observing the first frame. This is possible because we take object information into account as another essential context information. For instance, the starting position of the object makes it possible to anticipate a \textit{pick} action after observing only one image. 

Something similar occurs with \textit{receive} actions, where the object is usually out of the scene, being held by a volunteer. For such actions, after seeing the first frame, it is not possible to assure which is the action, once it depends on the direction. However, we can tell that it will be a \textit{receive} action. Therefore, the correct anticipation comes after perceiving the gaze or the movement direction. This helps us to eliminate less likely actions and allows us to focus on information that helps to find the right action. Fig.~\ref{fig:sample_ambuiguities_object} illustrates four situations where there are significant ambiguities between actions, and the object information is critical to reduce it.

 \subsection{Anticipation}

\begin{figure} [h!]
\centering
\subfloat[Place left]{\includegraphics[width =1.0\linewidth]{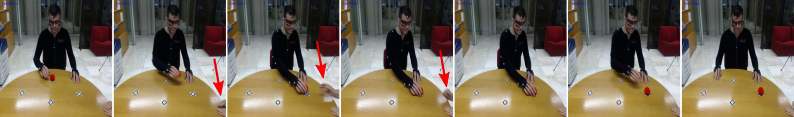}}\\
\subfloat[Give left]{\includegraphics[width =1.0\linewidth]{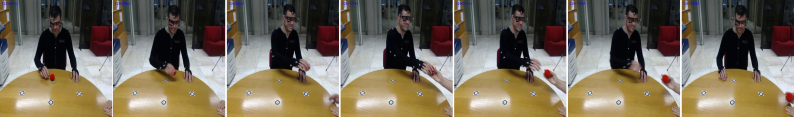}}
\caption{ Two action samples from the Acticipate dataset. In (a), the volunteer wrongly anticipated the action, thinking it would be a give left action (shown in (b)) instead of a place left.}
\label{fig:fooled}
\end{figure}

Although we are able to anticipate actions in the extended dataset, in some cases, there are issues about the anticipation that must be taken into account. Even people can have their prediction capacity compromised by overconfidence. In a particular case, as presented in Fig.~\ref{fig:fooled}, the volunteer wrongly anticipated an action after observing a movement similar to another one. Her confidence in her prediction deceived her. Thus, even people, in some situations, need to be more sure about the action before making a decision. If a person can be fooled by his/her overconfidence, this problem is possibly more significant in a computational model. 
 
The overconfidence about a prediction could lead the model to make a wrong decision in a real-time situation. This problem can be mitigated by providing the model with the ability to estimate the uncertainty about its prediction. A deterministic model, even with a high value of probability threshold (e.g., $p>0.9$), could wrongly anticipate an action when it is overconfident about its prediction. This overconfidence in prediction can be provoked by a lack of data to prevent the model from ambiguous classes. 
 
Thus, as mentioned in \citep{jain2015car}, a possible solution to increase the model certainty is to lead it to make more $z$ predictions before deciding on the correct action class. In this way, if the predicted class remains for the next $z$ observations, the model can be more confident about the correct class and can anticipate the action. However, even though it looks like a good solution, what is the best size for $z$? An inaccurate choice of this new parameter can postpone the anticipation of actions that have no ambiguity problem in $z$ observations. Additionally, $z$ may not be enough for actions with more ambiguities. 
 
 A better solution is to use as threshold an uncertainty value rather than a probability value. Thus, the model can anticipate an action when it is more certain about its prediction. Therefore, ambiguous actions, which likely provide more uncertainty to the model, would need more observations to be anticipated properly. On the other hand, those with less ambiguities could be anticipated previously. This solution can be taken as a tailored $z$ value for each action chosen by the model during training.

 \subsection{Hypotheses and contributions}
These observations led us to raise three main hypotheses regarding the Acticipate dataset:
\begin{enumerate}
\item 
more actions are likely to cause more ambiguities; 
\item 
context information can help to distinguish different actions represented by similar movements;
\item 
uncertainty is a more reliable and effective threshold to anticipate actions than probability values.
\end{enumerate}

In this work, the gaze and the object's position represent the context of each action. So, we propose a model based on Artificial Neural Networks (ANNs) that anticipates actions represented by sequences of data with varying lengths. The proposed model has two versions: a deterministic and a stochastic one.

 \section{Bayesian Neural Networks and Uncertainty}
\label{sec:Bayesian_networks}
 
Deep neural networks are usually trained by optimization algorithms based on Stochastic Gradient Descent (SGD). As SGD uses the gradient of the weights, it needs the loss function to be differentiable for all weights, which implies the weights must be deterministic variables. In consequence, most deep neural network models are deterministic, so they are unable to provide their uncertainty about their predictions. Thus, to measure uncertainty in this type of model, we can create a Bayesian Neural Network (BNN).

In a Bayesian model a posterior distribution must be inferred by applying the Bayes rule:
\begin{equation}
\label{eq:bayes_inference}
p(\THETA |\Dcj) = \frac{p(\Dcj | \THETA)p(\THETA)}{\int p(\Dcj | \THETA)p(\THETA) d\THETA},
\end{equation}
where $p(\THETA | \Dcj)$ is the posterior distribution over $\THETA$ after observing data $\Dcj$; 
$p(\Dcj | \THETA)$ is the likelihood of $\Dcj$; 
$p(\THETA)$ is the prior belief about the distribution of $\THETA$; 
and $\int p(\Dcj | \THETA)p(\THETA) d\THETA$ is the the normalization term (a.k.a evidence or marginal likelihood).

In many cases, the evidence term in Eq.~\eqref{eq:bayes_inference} turns the posterior inference intractable. However, some works attempted to solve this problem via Variational Inference (VI)~\citep{blei2017variational}. In 2011, \citep{graves2011practical} proposed in detail how to use VI in Bayesian Neural Networks so that a Gaussian distribution with known parameters could approximate its posterior distribution. Although effective, VI was not yet an easy task to accomplish. Therefore, in 2013, \citep{kingma2013VAE} proposed a way to train a BNN with VI thought a technique called reparametrization trick, which consists of drawing the activation $Z$ of a layer $l$ from a standard factorized Gaussian distribution. 

In this sense, the layer $l$ outputs two values, $\mu$ and $\sigma$, which represents the mean and variance of a factorized Gaussian distribution $N(\mu,\sigma)$, respectively. Next, the activation of $l$ is drawn from $Z \sim N(\mu,\sigma)$. Aiming to approximate $Z$ by a standard factorized Gaussian distribution ($N(0,1)$), the authors use variational inference. However, as $Z$ is now stochastic, SGD algorithms can not be used to train the parameters of $l$. To solve this problem, they proposed to parametrize $Z$ so that $\mu$ and $\sigma$ being deterministic with respect to $Z$, and, by consequence, differentiable with respect to a cost function. In that way, an SGD algorithm can be used to train the parameters of layer $l$. Eq.~\eqref{eq:repar_trick} presents this approach, so-called reparametrization trick.
 
\begin{eqnarray}
\label{eq:repar_trick}
Z &\sim& N(\mu,\sigma)\nonumber\\
Z &=& \mu + \epsilon\sigma . 
\end{eqnarray}

Here, the noise $\epsilon\sim N(0,1)$ is responsible for the stochasticity in $Z$.

Even with a significant contribution, the authors in \citep{kingma2013VAE} used $Z$ as the last layer of an encoder, not in all network activations or weights. Hence, in 2015, \cite{blundell2015weight} proposed to use this approach to create a BNN considering each weight as a distribution instead of a deterministic variable. The reparametrization trick allowed them to use the SGD algorithm to train the model, and to use VI to approximate the factorized weight posterior distribution to a distribution with known parameters. This approach is called Bayes By BackProp (BBB).

Other approaches, as MC dropout\citep{gal2016dropout} and Variational dropout\citep{kingma2015variational_dropout}, use dropout to obtain an approximation of a Bayesian model.
 
In MC dropout, the model must have a dropout function before each weight layer. Thus, the Bayesian approximation is achieved by randomly deactivating weights based on a Bernoulli distribution with the probability of $1-p$, where $p$ is a hyperparameter. The name MC dropout is given once the model prediction is calculated by the average of $S$ Monte Carlo (MC) samples on the model with the dropout enabled.

Variational Dropout uses the local reparametrization and VI to train and to approximate the neural network model of a Bayesian model. With the reparametrization trick in BBB ( Eq.~\eqref{eq:repar_trick}), after a layer $i$ receives $\xx_i$ as input, it first samples the weights $\THETA$ from a Gaussian distribution $N(\mu,\sigma)$ and then computes the activation $\hat{y} =  \THETA^T \xx $ as the inner product between $\xx$ and $\THETA$. On the other hand, in local reparametrization, the activations are sampled directly from a factorized Gaussian distribution, as shown in Eq.~\eqref{eq:local_reparametrization}:

\begin{eqnarray}
\label{eq:local_reparametrization}
\mu &=& \THETA^T \xx \nonumber\\
\sigma &=& (\THETA^2)^T \xx^2  \\
\hat{y} &\sim& N(\mu,\sigma) \nonumber.
\end{eqnarray}
\noindent where, $\mathbf{b}^2 = \mathbf{b} \circ \mathbf{b}$, where $\circ$ represents the pointwise multiplication operator.

This local reparametrization technique can be used in conjunction with a noise $\xi \sim N(1,\alpha)$ in order to get the posterior $p(\omega | D) = N(\theta,\alpha \THETA^2)$, where $\omega$ is the variational parameter, $\THETA$ is the model weight and $\alpha= p/(1 - p)$. Eq.~\eqref{eq:variational_dropout} presents the variational dropout approach.

\begin{equation}
\label{eq:variational_dropout}
\hat{y} = \THETA^T(\xx \circ \xi).
\end{equation}

As $\xi$ is drawn from a Gaussian distribution, the marginal distribution $\hat{y} = p(\hat{y}|\xx)$ is also a Gaussian distribution. Thus, one can sample $\hat{y}$ directly from its marginal distribution $p(\hat{y}|\xx)$, as presented in Eq.~\eqref{eq:var_drop_marginal}.

\begin{eqnarray}
\label{eq:var_drop_marginal}
\mu &=& \THETA^T \xx \nonumber\\
\sigma &=& \alpha (\THETA^2)^T \xx^2\\
\hat{y} &\sim& N(\mu,\sigma)\nonumber.
\end{eqnarray}

Even though $p$ in MC dropout is a hyperparameter, $\alpha$ in variational dropout can be taken as a trained parameter, giving different importance for each element in $(\THETA^2)^T \xx^2$. 

In a Bayesian model, regardless of the particular approach employed to infer the posterior distribution, the prediction of an observation $\xx^*$ is calculated by integrating the likelihood of $\xx^*$ over the entire posterior distribution (Eq.~\eqref{eq:bayes_prediction}). As this process involves an intractable integration, an unbiased approximation can be obtained by a Monte Carlo simulation, as presented in Eq.~\eqref{eq:bayes_prediction_MC}.

\begin{eqnarray}
\label{eq:bayes_prediction}
p(y^* |\xx^*) &=& \int p(D|\THETA)p(\THETA |D)d\THETA \\
\label{eq:bayes_prediction_MC}
&\approx& \frac{1}{S}\sum_{s=1}^S p(y^*|\xx^*;\theta_s).
\end{eqnarray}

\noindent Here, $S$ is the number of samples, $y^*$ is the probability distribution of classes given $\xx^*$, and $\theta_s \sim p(\theta | D) $ is the $s^{th}$ parameter $\theta$ drawn from the posterior $p(\theta | D)$. For the variational dropout model, this posterior is $p(\omega|D)$. However, for MC dropout, this posterior distribution is represented by the dropout function inside each network layer.

\subsubsection{Uncertainty}

There are two main types of uncertainty in Bayesian modeling: aleatoric and epistemic. 
Aleatoric is the uncertainty of an event (a.k.a irreducible uncertainty). In a classification problem, this uncertainty is related to the event that generates the data. Therefore, even though some works propose ways to assess the aleatoric uncertainty of a model \citep{gal2017uncertainties,hafner2018ccontrastive_np}, it is not an easy task to perform, as in most cases, one can not know how the data was sampled or which event generated them. 

Epistemic uncertainty assesses the model uncertainty about the data and can be easily calculated when the model is stochastic. This type of uncertainty can be decreased by observing more data. Thus, it is important when one wants to know which class needs more data to improve model prediction. A detailed explanation about uncertainties for Bayesian Deep Neural Networks can be find in \citep{gal2016thesis}.

In this work, we are interested in determining the uncertainty of the model's prediction, which corresponds to its epistemic uncertainty. In this sense, the more data it receives during training, the more confident it would be about its predictions. Thereby, actions with fewer samples data would lead the model to uncertain predictions. In this case, it is possible to use epistemic uncertainty to realize when the model should wait for more observations to increase its certainty about prediction, and then anticipate the action correctly.

The epistemic uncertainty of a Bayesian Neural Network model can be estimated by the entropy or the mutual information metrics~\citep{gal2016thesis}. In the case of an MC simulation with $S$ samples, a model with $C$ actions can calculate the entropy of theses predictions (samples) by using Eq.~\eqref{eq:entropy} and the mutual information by Eq.~\eqref{eq:mutual_information}. 
\begin{equation}
\mathbb{E}_{pred}(x,c) = \frac{1}{S} \sum_{s=1}^S p(y = c|x;\theta_s),
\label{eq:entropy}
\end{equation}

\begin{equation}
\mathbb{H}(x) = - \sum_{c=1}^C  \mathbb{E}_{pred}(x,c)\textrm{ log}(\mathbb{E}_{pred}(x,c)),
\label{eq:entropy}
\end{equation}

\begin{equation}
\mathbb{I}(x) = \mathbb{H}(x) +  \frac{1}{S} \sum_{c=1}^C \sum_{t=1}^S p(y = c|x;\theta_s) \,\textrm{log} \, p(y = c|x;\theta_s).
\label{eq:mutual_information}
\end{equation}

\section{Proposal}
\label{sec:proposal}

In this section, we present our proposed architecture, which is divided into four main steps:
(\textit{i}) feature extraction and selection, 
(\textit{ii}) feature embedding,
(\textit{iii}) classification model, and 
(\textit{iv}) a novel decision-making criterion.
The next topics will cover each step in detail. 
  
\subsection{Feature extraction and selection}

This work proposes to use gaze and object position as the context information, and the evolution of the 2D body joints features as the movement information to perform action anticipation. Our approach aims to use only RGB images, where gaze and skeleton joints information are not straight available.

Therefore, to obtain the gaze and skeleton joints of the people present in the images, we consider to use the Openpose model\citep{cao2018openpose} over each RGB image to extract such information. We used the Openpose version trained for COCO dataset that provides 19 2D joints for the body and 25 2D points for each hand.
  
It is important to mention that in \cite{schydlo2018anticipation}, the authors used gaze and 3D body joints since they had glasses and a MoCap system, while, in our approach, we have only 2D joints to use as data, because we are considering just RGB images. Also, because the actor wore glasses during data acquisition, algorithms for 2D gaze estimation did not work. For this reason, we decided to use the head joints as information that likely may represent head direction or even gaze. However, this representation is a task to be assumed by the model. Aiming to reduce dimensionality, we calculated the central point of each hand instead of using their 25 2D points directly.

\begin{figure} [h!]
\centering
\includegraphics[width =0.9\linewidth]{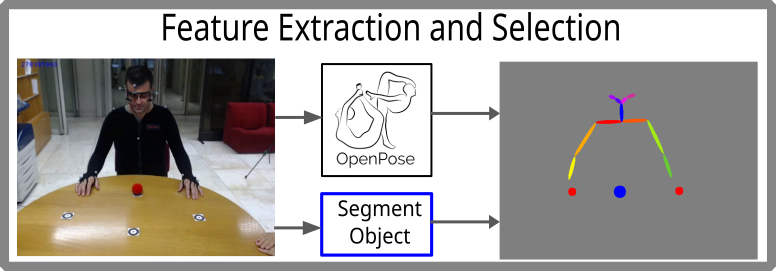}
\caption{Summary of the feature extraction and selection step.}
\label{fig:feature_extraction}
\end{figure}

For the object information, we extracted the central point of the red ball for each frame using a segmentation method. This pre-processing procedure is summarized in Fig.~\ref{fig:feature_extraction} and described as follows:
  
\begin{enumerate}
\item 
Openpose model receives an RGB image representing an observation. This operation results in 19 joints and 25 hand points from each user present in the image. 
\item 
A filter to remove false-positive users is applied.
\item 
Select the most important joints (arms, shoulders, and head)
\item 
Use hand points to calculate the central point of each hand
\item 
Give the same RGB image as input to segmentation function to extract the central point of the object.
\end{enumerate}
      
\subsection{ Feature Embedding}
\label{embedding}

After the pre-processing step, head information is represented by: 
(\textit{i}) five 2D points ($\mathbf{v}_h \in \mathbb{R}^{10 \times 1}$); 
(\textit{ii}) object information by one 2D point ($\mathbf{v}_o \in \mathbb{R}^{2 \times 1}$); and (\textit{iii}) user pose (movement) by nine 2D points ($\mathbf{v}_m \in \mathbb{R}^{18 \times 1}$), where the first seven points represent arms and shoulders, and the last two points represent the hands. Notice that, movement, head, and object have different quantity of points, which generate an unbalanced feature vector. Because of that, the model may consider the movement more important than the other features. Therefore, we propose to balance the input source by using an embedding structure, in such a way that movement and context features have the same dimension. Besides that, to represent the context, head and object features were also defined with the same dimension, so they had the same importance. The embedding process is explained below.

\begin{enumerate}
\item 
Embed head information: $\mathbf{e}_h = f(\mathbf{W}^T_h\mathbf{v}_h+\mathbf{b}_h$), where $\mathbf{e}_h \in \mathbb{R}^{16\times1}$, $\mathbf{W}_h\in \mathbb{R}^{10\times16}$ and $\mathbf{b}_h\in \mathbb{R}^{16\times1}$.
\item 
Embed object information: $\mathbf{e}_o = f(\mathbf{W}^T_o\mathbf{v}_o+\mathbf{b}_o$), where $\mathbf{e}_o \in \mathbb{R}^{16\times1}$, $\mathbf{W}_o\in \mathbb{R}^{2\times16}$ and $\mathbf{b}_o\in \mathbb{R}^{16\times1}$.
\item 
Embed movement information: $\mathbf{e}_m = f(\mathbf{W}^T_m\mathbf{v}_m+\mathbf{b}_m)$, where $\mathbf{e}_m \in \mathbb{R}^{16\times1}$, $\mathbf{W}_m\in \mathbb{R}^{18\times16}$ and $\mathbf{b}_m\in \mathbb{R}^{16\times1}$.
\item 
Embed context information: $\mathbf{e}_c = f(\mathbf{W}^T_c[\mathbf{e}_o|\mathbf{e}_h] + \mathbf{b}_c)$, where $\mathbf{e}_c \in \mathbb{R}^{16 \times 1}$, $\mathbf{W}_c \in \mathbb{R}^{32\times16}$, and $\mathbf{b}_c\in \mathbb{R}^{16\times1}$ and $|$ is a vector concatenation operator.
\item 
Create the embedded input vector: $\mathbf{e}_{cm} = [\mathbf{e}_c | \mathbf{e}_m]$, where $\mathbf{e}_{cm} \in \mathbb{R}^{32\times1}$.
\end{enumerate}
  
    \begin{figure} [h!]
    \centering
    \includegraphics[width =0.7\linewidth]{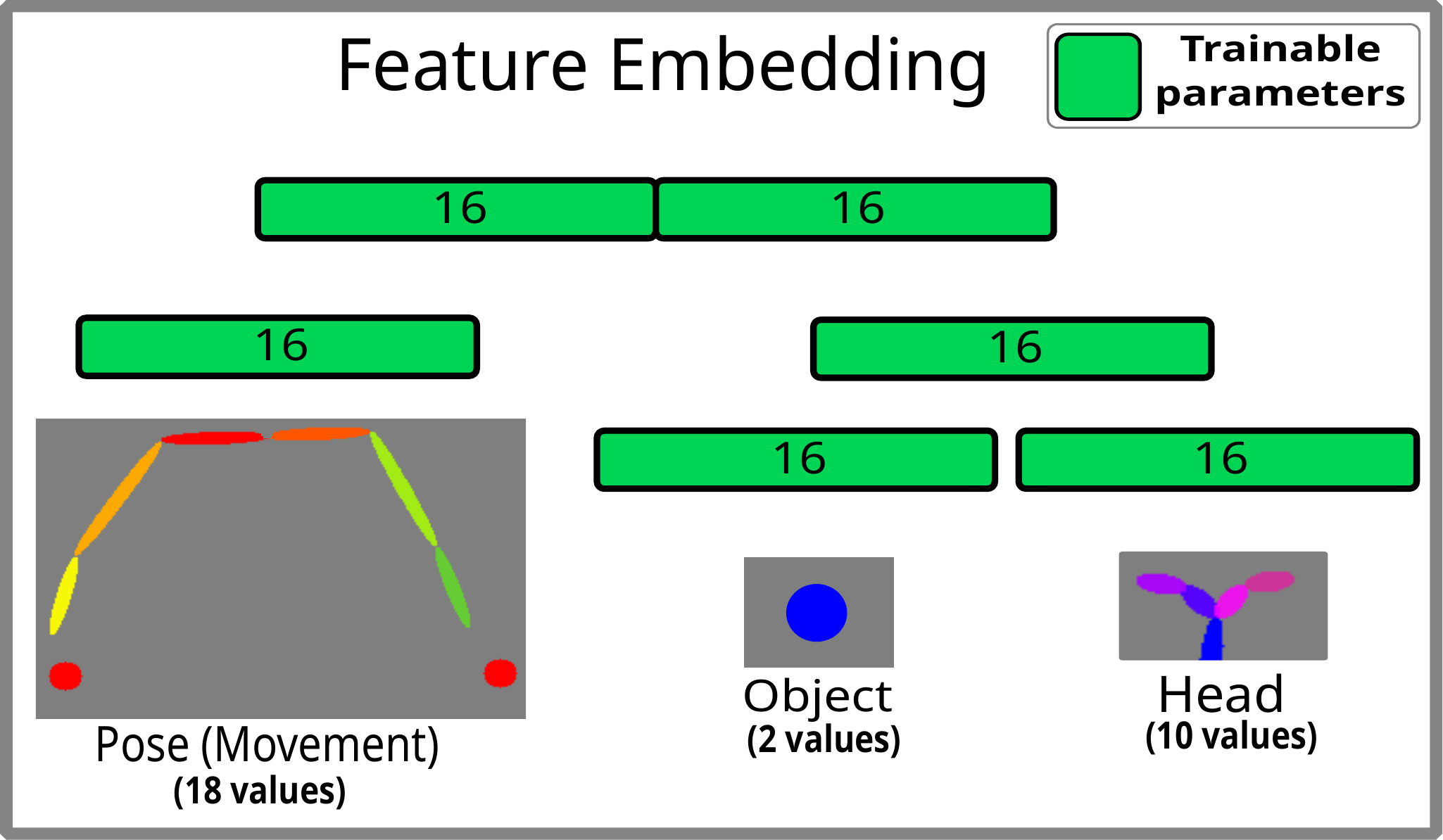}
    \caption{Feature embedding process for each observation. The connections between joints as well as the object shape are showed just for the sake of visualization. However, only their points are used.}
    \label{fig:embedding}
    \end{figure}
  
Each $\mathbf{W}_*$ and $\mathbf{b}_*$ represents weights that are trained by the model, whereas $f(\bullet)$ represents the ReLU activation function. Thus, during the training phase, the model simultaneously learns to incorporate observations and classify actions.

\subsection{Classification model}

Since the problem addressed here has a sequential nature, and we assume that there exist dependencies between observations of different timesteps, we can treat the problem in two ways: considering that all sequences are limited to a fixed size of $M$ observations or assuming the original size of sequences.

\begin{figure} [h!]
\centering
\includegraphics[width =0.7\linewidth]{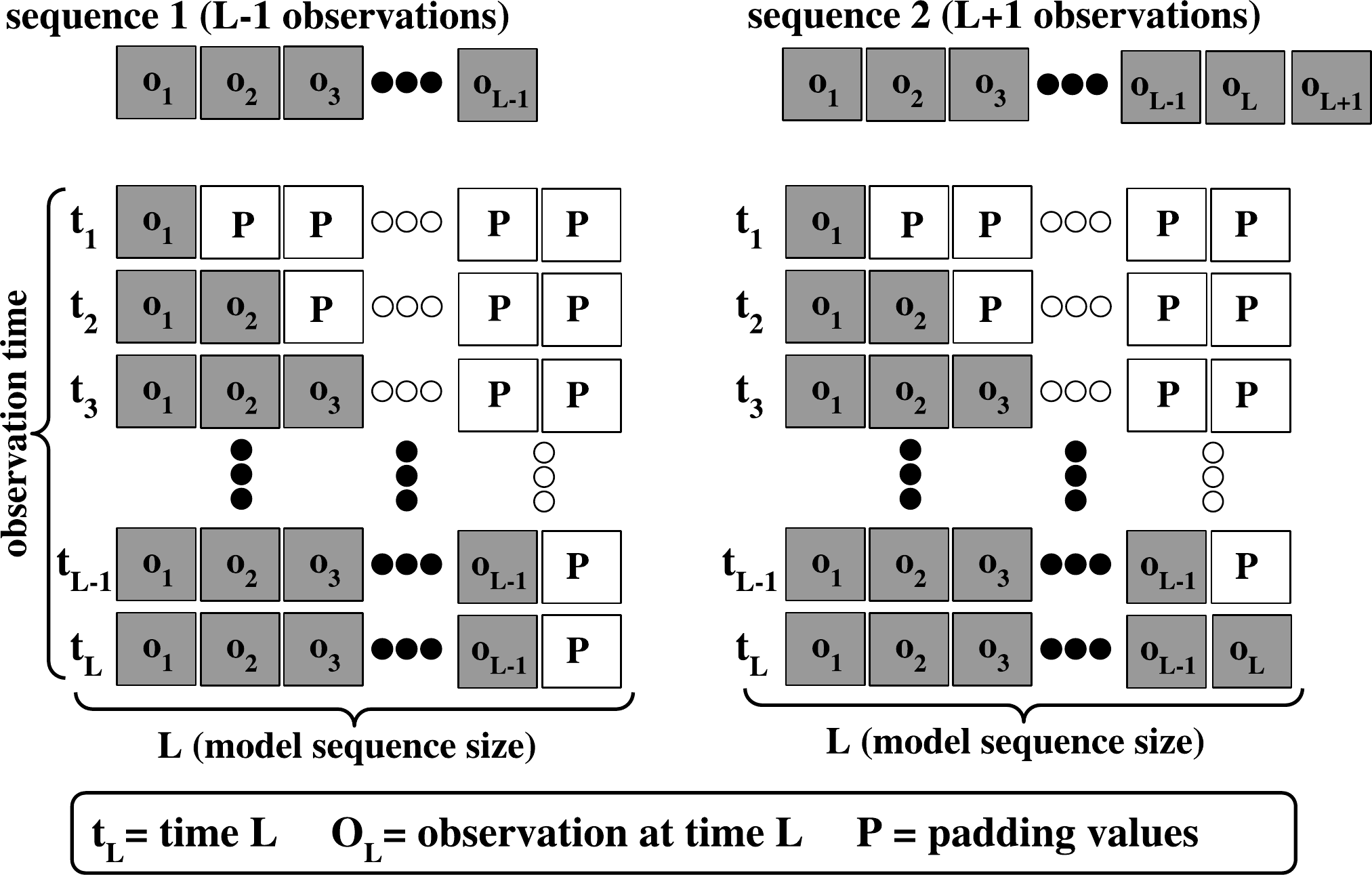}
\caption{Representation of two sequences with different sizes for a model that receives a fixed input sequence of size M. At each time $t$, the $t^{th}$ observation from the sequence is added to the fixed model sequence. Therefore, the model can predict the action represented by the $t$ observations.}
\label{fig:fixed_size_sequence}
\end{figure}

The problem with the first approach is to disregard the variance in the size of all sequences, besides introducing a new hyperparameter $M$. Thus, a sequence with $T$ observations ($T>M$) must be truncated at observation $T-M$, meanwhile a sequence with $L$ observations ($L<M$) must be padded with $M-L$ values (an illustration of this process can be seen in Fig.~\ref{fig:fixed_size_sequence}). Therefore, aiming to acquire a score that corresponds to the chance of a sequence up to time $t$ ($t\leq M$) belongs to one action class, we can feed the model with a sequence of size $M$, where the first $t$ observations came from the real sequence, and the last $M-t$ are padding values. An advantage of this approach is to enable the use of non-sequential models, like Naive Bayes or Multilayer Perceptron, to classify the sequence, since the dependence between observations can be treated as dependence between features. The main problem with this approach is the waste of processing power when the sequence is starting (since the majority of the input data is padded with default values), besides the possible poor performance when the dependence between long sequences must be considered.

For the second case, only a sequential model can be used, as the size of the sequence is not available. In this case, models as HMM (Hidden Markov Model) and CRF (Conditional Random Fields) are possible candidates. However, these models assume the Markovian condition: a given observation depends only on the previous one. This assumption might not capture long dependencies on a sequence, which occurs during action execution. Therefore, we decided to use LSTM (Long-Short Term Memory)\citep{hochreiter1997lstm}, a variant of RNN that can capture long dependencies in a sequence of observations. 
  
An LSTM contains four trainable gates. These gates are responsible for capturing long and short dependencies in a sequence. An LSTM cell receives as input an observation vector, a hidden state, and an echo cell. The input vector represents the actual observation; the hidden state represents the short-term memory and chooses what information should be paid attention in the next observation. The echo cell represents the long-term memory. At each new observation, the echo cell stores important pieces of information about the actual observation and forget part of its past when it considers less significant. 

LSTM has been used mainly for NLP \citep{vaswani2017attention,devlin2019bert} but in the last few years recognition tasks in videos are commonly using it as well. The Eq.~\eqref{eq:forget_gate}-\eqref{eq:hidden_state} represent all LSTM gates and activations, respectively: 

\begin{itemize}
\item 
Forget gate, $\mathbf{f}_t$, forget part of the memory stored in the echo cell.
\begin{equation}
\label{eq:forget_gate}
\mathbf{f}_t = \sigma_g(\mathbf{W}_{f} \mathbf{x}_t + \mathbf{U}_{f} \mathbf{h}_{t-1} + \mathbf{b}_f). 
\end{equation}
\item 
Input gate, $\mathbf{i}_t$, select part of the observation to be stored into the next echo cell.
\begin{equation}
\label{eq:input_gate}
\mathbf{i}_t = \sigma_g(\mathbf{W}_{i} \mathbf{x}_t + \mathbf{U}_{i} \mathbf{h}_{t-1} + \mathbf{b}_i).
\end{equation}
\item 
Output cell, $\mathbf{o}_t$, select what part of the input will be propagated to the next observation by the hidden state.
\begin{equation}
\label{eq:output_gate}
\mathbf{o}_t = \sigma_g(\mathbf{W}_{o} \mathbf{x}_t + \mathbf{U}_{o} \mathbf{h}_{t-1} + \mathbf{b}_o).
\end{equation}
\item 
Update gate, $\mathbf{g}_t$, normalize the observation in order to store it into the next echo cell. Part of this information will be forgot by using the input gate.
\begin{equation}
\label{eq:update_gate}
\mathbf{g}_t = \tanh(\mathbf{W}_{o} \mathbf{x}_t + \mathbf{U}_{o} \mathbf{h}_{t-1} + \mathbf{b}_o).
\end{equation}
\item 
Next echo cell, $\mathbf{c}_t$, forget part of the past observations and store part of the new one.
\begin{equation}
\label{eq:echo_cell}
\mathbf{c}_t = f_t \circ \mathbf{c}_{t-1} + \mathbf{i}_t \circ \mathbf{g}_t.
\end{equation}
\item 
next hidden state, $\mathbf{h}_t$, select a part of the normalized echo cell by using the output gate.
\begin{equation}
\label{eq:hidden_state}
\mathbf{h}_t = \mathbf{o}_t \circ \tanh(\mathbf{c}_t).
\end{equation}
\end{itemize}
  
where $\sigma_{g}(\bullet)$ and $\tanh(\bullet)$ are, respectively, \textit{sigmoid} and \textit{hyperbolic-tangent} activation functions.
  
\begin{figure*}[h!]
\centering
\includegraphics[width =1.0\linewidth]{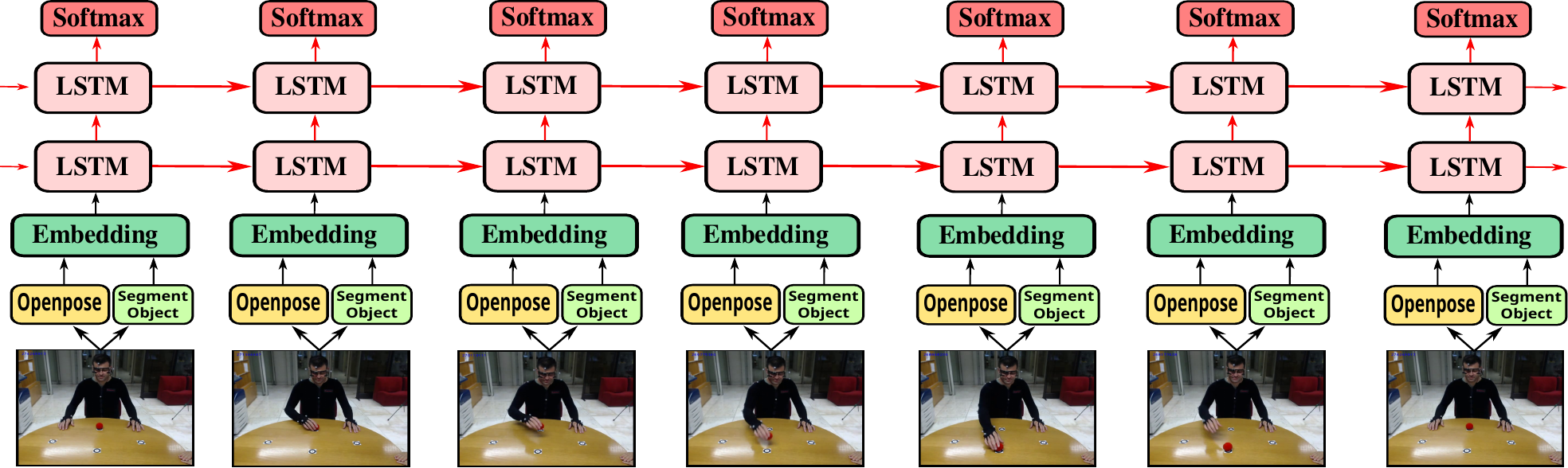}
\caption{Proposed model architecture}
\label{fig:model}
\end{figure*}
      
Fig.~\ref{fig:model} presents the proposed model. It comprises two LSTM layers followed by a softmax classifier. 
The first LSTM cell receives as input at timestep $t$ the embedded input vector $\mathbf{e}^{(t)}_{cm}$, the hidden state $\mathbf{h}^{(t)}_1 \in \mathbb{R}^{64 \times 1}$ and the echo cell $\mathbf{c}^{(t)}_1 \in \mathbb{R}^{64 \times 1}$. 
The second LSTM cell receives as input the hidden state $\mathbf{h}^{(t)}_1$ resulted from the first layer, the hidden state $\mathbf{h}^{(t)}_2 \in \mathbb{R}^{64 \times 1}$, and the echo cell $\mathbf{c}^{(t)}_2 \in \mathbb{R}^{64 \times 1}$. 
Next, a fully connected layer receives as input $\mathbf{h}^{(t)}_2$, applies a transformation using a matrix $\mathbf{W}_{fc} \in \mathbb{R}^{d \times 64}$ ($d$ is the number of actions) and normalize it using a \textit{softmax} function. So, 

$$\hat{\ssx}^{(t)} = \text{softmax}(\mathbf{W}_{fc}\mathbf{h}^{(t)}_2),$$

\noindent where $\hat{\ssx}^{(t)} \in \mathbb{R}^{d \times 1}$, and each element of $\hat{\ssx}^{(t)}$ can be interpreted as the probability of an action given the embedded input vector in the timestep $t$. Now, With this result and choosing a probability threshold value $p$, by using Eq.~\eqref{eq:anticipation_threshold}, the anticipation can be accomplished.

\subsection{Decision-making criterion}   

As mentioned before, this work proposes a novel decision-making criterion based on the model uncertainty. In this sense, because the proposed model is deterministic, we propose to use three new stochastic versions of it: A Bayesian LSTM using Bayes By BackProp ($\textrm{BLSTM}_{BBB}$), an MC dropout Bayesian LSTM based on \citep{gal2016rnn} ($\textrm{BLSTM}_{MC}$), and a Variational Dropout Bayesian LSTM ($\textrm{BLSTM}_{VD}$). 
  
With these stochastic models, the uncertainty is obtained by running the architecture of Fig.~\ref{fig:model} $S$ times (MC sampling) using the same input sequence. As the models are stochastics, they must give a different value for each prediction. Thus, the Mutual Information (MI) over the $S$ predictions give us the epistemic model uncertainty about the class prediction for the respective observation. MI is calculated using Eq.~\eqref{eq:mutual_information}. Therefore, we propose to use a threshold over the mutual information (our new decision-making criterion) to anticipate actions. For this, Eq.~\eqref{eq:anticipation_threshold} must be redefined as:
  
\begin{equation}
\label{eq:uncert_ant_func}
g(\hat{\ssx}^{(t)} ,u) = \PARTCTWO{\ARGMAXL(m( \hat{\ssx}^{(t)} ))}{h(\hat{\ssx}^{(t)} ) < u}{-1}{\OTHERWISE},
\end{equation} 

\noindent where $u$ is an uncertainty value, $h$ is the mutual information function (Eq.~\eqref{eq:mutual_information}), and $m$ is the average of the $S$ predictions.
  
Even though one can use the entropy (Eq.~\eqref{eq:entropy}) to measure the uncertainty, we chose to use MI because it takes into account not only entropy between classes (averaged over the $S$ predictions) but also the mean entropy between them all.

\section{Experiments}
\label{sec:experiments}

As discussed in Sec.~\ref{sec:intuitions}, all experiments in this work used the Acticipate dataset. From the dataset, we extracted four different kinds of data by using the procedure described in Sec.~\ref{sec:proposal}: head points, object position, arm joints, and hand position. The head points and object position forms the context; arm joints and hand positions form a pose, which evolution in time represents the movement (also called here main entity). To better compare results and reach more reliable conclusions about how each source of information influences the action anticipation, we decided to carry out experiments using different combinations of the context (head and object) and movement, for the original version of the dataset (6 actions) and its extended version (12 actions). 

The most common approaches, presented in Sec.~\ref{sec:related_works}, need a large dataset to be trained. Hence, they are not suitable to be used here, with the Acticipate dataset. However, to better compare and discuss our results, we used as baselines the proposal in \cite{schydlo2018anticipation}, that uses Acticipate dataset, in addition to five classical models: Naive Bayes (NB), Multilayer Perceptron (MLP), 1D Convolutional Neural Network (CONV-1D), Support Vector Machine (SVM) and HMM. For the first four models, as mentioned before (Sec. \ref{sec:proposal}), we used a fixed sequence size (as illustrated  in Fig.~\ref{fig:fixed_size_sequence}), while for HMM we used the sequences with their original sizes.

Considering the architecture of the baseline models:
\begin{itemize}
\item 
NB uses a Gaussian Model to predict its conditional probability;
\item 
MLP is composed of only one hidden layer;
\item 
CONV-1D has three stacked 1D Convolutional layers followed by an MLP as classifier; 
\item 
SVM implements its non-linear version by using an RBF kernel; and 
\item 
HMM has its emissivity probability drown from a Gaussian Distribution, which enables continuous observations.
\end{itemize} 

The five models were trained with the original dataset (6 actions) and its extended version (12 actions). For the experiments with these five models, we did not apply the proposed embedding technique. So each observation is represented by a vector with 18 values corresponding to the arms, two values for the object, and 10 for head points (see Sec.~\ref{embedding}). By combining these features (movement, object, and head), we carried out a total of 45 different baseline experiments. However, for the sake of explanations, we will present only the most conclusive ones. We also carried out more 11 experiments with different versions of the proposed model, which will be explained next. Table \ref{tab:experiment_list} summarizes the main 16 experiments.

\begin{table*}[]
\centering
\small
\caption{List of all experiments}
\label{tab:experiment_list}
\begin{tabular}{lllccccc}
\hline
\multirow{2}{*}{Id} & \multicolumn{1}{c}{\multirow{2}{*}{Model}} & \multicolumn{1}{c}{\multirow{2}{*}{Type}} & \multirow{2}{*}{Movement} & \multicolumn{2}{c}{Context} & \multicolumn{2}{c}{Dataset Version} \\
  & \multicolumn{1}{c}{} & & & Head  & Object    & 6 actions  & 12 actions    \\\hline\hline 
1 & $NB$   & baseline     & \v & \v &   & \v &   \\
2 & CONV-1D  & baseline & \v & \v &   & \v &   \\
3 & $MLP$  & baseline     & \v & \v &   & \v &   \\
4 & $SVM$  & baseline     & \v & \v &   & \v &   \\
5 & $HMM$  & baseline     & \v & \v &   & \v &   \\\hline
6 & $DLSTM_{6m}$   & Deterministic    & \v &  &   & \v &   \\
7 & $DLSTM_{6mh}$  & Deterministic    & \v & \v &   & \v &   \\\hline
8 & $DLSTM_{12m}$  & Deterministic    & \v &  &   &   & \v \\
9 & $DLSTM_{12h}$  & Deterministic    &   & \v &   &   & \v \\
10 & $DLSTM_{12o}$  & Deterministic    &   &  & \v &   & \v \\
11 & $DLSTM_{12mh}$  & Deterministic    & \v & \v &   &   & \v \\
12 & $DLSTM_{12mo}$  & Deterministic    & \v &  & \v &   & \v \\
13 & $DLSTM_{12mho}$ & Deterministic    & \v & \v & \v &   & \v \\ \hline
14 & $BLSTM_{MC}$   & MC Dropout     & \v & \v & \v &   & \v \\
15 & $BLSTM_{VD}$   & Variational Dropout & \v & \v & \v &   & \v \\
16 & $BLSTM_{BBB}$  & Bayes By BackProp  & \v & \v & \v &   & \v \\ \hline 
\end{tabular}
\end{table*}

The experiments $6$ and $7$ (original dataset) in Table~\ref{tab:experiment_list} provide results that can be compared with \citep{schydlo2018anticipation} and the baseline experiments (5 first experiments), which will validate our proposal against the other models. The experiments from $8$ to $13$ provide results to show the ambiguities between actions and the importance of context to anticipate them. The last three experiments show the importance of uncertainty in an anticipation model, and the contribution of the proposed decision-making criterion based on the uncertainty. For this reason, three Bayesian models were implemented: MC dropout, Variational Dropout, and Bayes by Backprop. For variational dropout, as mentioned before, we opt to use $\alpha$ (Eq.~\eqref{eq:var_drop_marginal}) as a trainable parameter. With these three models, we can identify which model is best for this kind of application.

\subsection{Experiment setups}

For each RGB image in the video, the pre-processing procedure described in Sec.~\ref{sec:proposal} was applied. Every missing data related to joints, hands and the object position were set to -1.  Additionally the padding value used in the fixed sequence size for the four first experiments was also considered -1. To evaluate the model's quality, the dataset was divided into 80\% for training and 20\% for testing. In each experiment, a 10-fold cross-validation over the training set was performed, where nine folds were used to train and one fold to validate. For each fold (round), the training process was finished when the recognition accuracy (the accuracy achieved at the last observation of the sequence) over the nine training folders in the previous five epochs was higher than 98\% (early stop) or when the iterations exceeded the maximum number of epochs.

The 16 experiments in Table \ref{tab:experiment_list} can be divided into four main categories: baselines with 6 actions (from 1 to 5), deterministic with 6 actions (6 and 7), deterministic with 12 actions (from 8 to 13) and stochastic with 12 actions (14,15 and 16). In both deterministic experiments, different configurations of the input data (movement, head and object) were achieved by assigning $-1$ to $\mathbf{v}_m$, $\mathbf{v}_o$, and/or $\mathbf{v}_h$ in the entire dataset. For instance, by assigning $-1$ to $\mathbf{v}_o$, the model considers only movement ($\mathbf{v}_m$) and head ($\mathbf{v}_h$) information. For each one of the 16 experiments, proper hyperparameters were chosen by using a Bayesian Optimization process. The hyperparameters used in the tests are presented in Table~\ref{tab:hyperparameters}~and~\ref{tab:baseline_hyperparameters}.

\begin{table*}[]
\small
\centering
\caption{List of hyperparameters used in each experiment with the proposed models}
\label{tab:hyperparameters}
\begin{tabular}{lcccc}
\hline
\multirow{2}{*}{Hyperparameters} & \multicolumn{4}{c}{Model Configurations} \\
                 & $DLSTM_*$  & $BLSTM_{MC}$ & $BLSTM_{VD}$ & $BLSTM_{BBB}$ \\\hline\hline
Batch size          & 216 & 216 & 216 & 32  \\
Truncate sequence       & 100 & 100 & 100 & 128 \\
Sequence size         & 100 & 100 & 100 & 64 \\
Max epochs          & 100 & 100 & 100 & 200 \\
LR              & 1e-2 & 1e-2 & 1e-2 & 1e-2\\
LR decay (per epoch)     & 1\% &1\% &1\% & 1\%   \\
Weight decay         & 1e-5 & 1e-5 & -- & -- \\
Dropout (keep prob)      & 0.7 & 0.2 & -- & --  \\
Optimizer           & Adam &Adam &Adam & Adam \\\hline                                                
\end{tabular}
\end{table*}

\begin{table*}[]
\small
\caption{List of hyperparameters used in each baseline experiment}
\label{tab:baseline_hyperparameters}
\small
\centering
\begin{tabular}{@{}lccccc@{}}
\hline
Hyperparameters & NB & MLP & SVM & CONV-1D & HMM \\ \hline\hline
Architecture & Gaussian & \begin{tabular}[c]{@{}c@{}}hidden layer\\ (48)\end{tabular} & \begin{tabular}[c]{@{}c@{}}Kernel\\ (RBF)\end{tabular} & \begin{tabular}[c]{@{}c@{}}Conv Layers\\ (1x7x32, 1x5x32, 1x5x32)\\ Hiddens layer\\ (64)\end{tabular} & \begin{tabular}[c]{@{}c@{}}States\\ (5)\end{tabular} \\ 
Sequence Size & 72 & 64 & 64 & 96 & 64 \\
Max Epoch & - & 50 & - & 50 & - \\
Batch Size & - & 32 &  & 32 & - \\
Learning Rate & - & 1e-3 & - & 1e-3 & - \\
Optimizer & - & Adam & - & Adam & - \\ \hline
\end{tabular}
\end{table*}

\subsection{Software and hardware environments}

The models, including Openpose (a deep neural network), were implemented in Pytorch v1.0 and scikit-learn v0.22. The Bayesian Optimization was implemented using the library Hyperopt\footnote{https://github.com/hyperopt/hyperopt}. Additional parts, as object segmentation, filters, and chart plot scripts, were implemented using OpenCV v4.1, Python v3.7, Numpy v1.16.4, and Matplotlib v2.2.3. The computer used in the experiments had the following configuration:

\begin{itemize}
\item Linux Operating System, distribution Ubuntu Server 16.04;
\item Intel Core i7-7700 processor, 3.60 GHz with four physical cores;
\item $32$ GB of RAM;
\item 1 TB of storage unit (hard drive);
\item Nvidia Titan V graphic card.
\end{itemize}

\section{Results and Discussions}
\label{sec:results}

For each one of the 16 experiments, after running the Bayesian Optimization over the 10-fold cross-validation, the best hyperparameter configuration found was used to train the model with the complete training set. Then, each model was tested with the test set, which was never seen by the model during training. The results over the test set were plotted in many charts and carefully analyzed. Some of these charts are presented and discussed in this section.

\subsection{Baseline models in the original dataset}

\begin{table*}[]

\caption{Comparing the results obtained by the models (baselines and DLSTM) at different observation ratios. The results for \citep{schydlo2018anticipation} were provided by the authors. The subscription $6mh$ indicates that the correspondent model was trained with the original dataset (6 actions) using movement (m) and head (h) as source of information.}
\label{tab:acc_baselines}
\tiny
\begin{tabular}{lllllllllll}
\hline
\multicolumn{1}{c}{\multirow{2}{*}{Models}} & \multicolumn{10}{c}{Percentage of observation} \\
\multicolumn{1}{c}{} & 10\% & 20\% & 30\% & 40\% & 50\% & 60\% & 70\% & 80\% & 90\% & 100\% \\ \hline\hline
$HMM_{6mh}$ & \textbf{19.05\%} & 19.05\% & 38.10\% & 57.14\% & 90.48\% & 95.24\% & \textbf{100.00\%} & \textbf{100.00\%} & \textbf{100.00\%} & \textbf{100.00\%}\\
$NB_{6mh}$ & \textbf{19.05\%} & 9.52\% & 9.52\% & 9.52\% & 19.05\% & 33.33\% & 71.43\% & 95.24\% & 95.24\% & 95.24\%\\
$CONV_{6mh}$ & 9.52\% & 14.29\% & 19.05\% & 19.05\% & 33.33\% & 47.62\% & 42.86\% & 61.90\% & 61.90\% & 80.95\%\\
$MLP_{6mh}$ & \textbf{19.05\%} & 19.05\% & 28.57\% & 28.57\% & 33.33\% & 47.62\% & 71.43\% & 90.48\% & 85.71\% & 95.24\%\\
$SVM_{6mh}$ & \textbf{19.05\%} & 19.05\% & 19.05\% & 19.05\% & 28.57\% & 71.43\% & 95.24\% & 95.24\% & 95.24\% & 95.24\%\\
$DLSTM_{6m}$ & 14.29\% & 28.57\% & 38.10\% & 71.43\% & 85.71\% & 95.24\% & 95.24\% & 95.24\% & 95.24\% & 95.24\%\\
$DLSTM_{6mh}$ & 14.29\% & \textbf{38.10\%} & \textbf{47.62\%} & \textbf{90.48\%} & \textbf{95.24\%} & \textbf{100.00\%} & \textbf{100.00\%} &\textbf{ 100.00\%} &\textbf{ 100.00\%} & \textbf{100.00\%}\\
 \citep{schydlo2018anticipation} (Pose 3D ) &  16.25\% & 21.25\% & 28.75\% & 51.25\% & 76.25\% & 85.00\% & 86.25\% & 86.25\% & 86.25\% & 85.00\%\\
 \citep{schydlo2018anticipation} (Pose 3D + Gaze) & 15.00\% & 16.25\% & 40.00\% & 73.75\% & 86.25\% & 97.50\% & 96.25\% & 92.50\% & 91.25\% & 87.50\%\\
 \\ \hline
\end{tabular}
\end{table*}

\begin{figure*} [h!]
\centering
\includegraphics[width =1\linewidth]{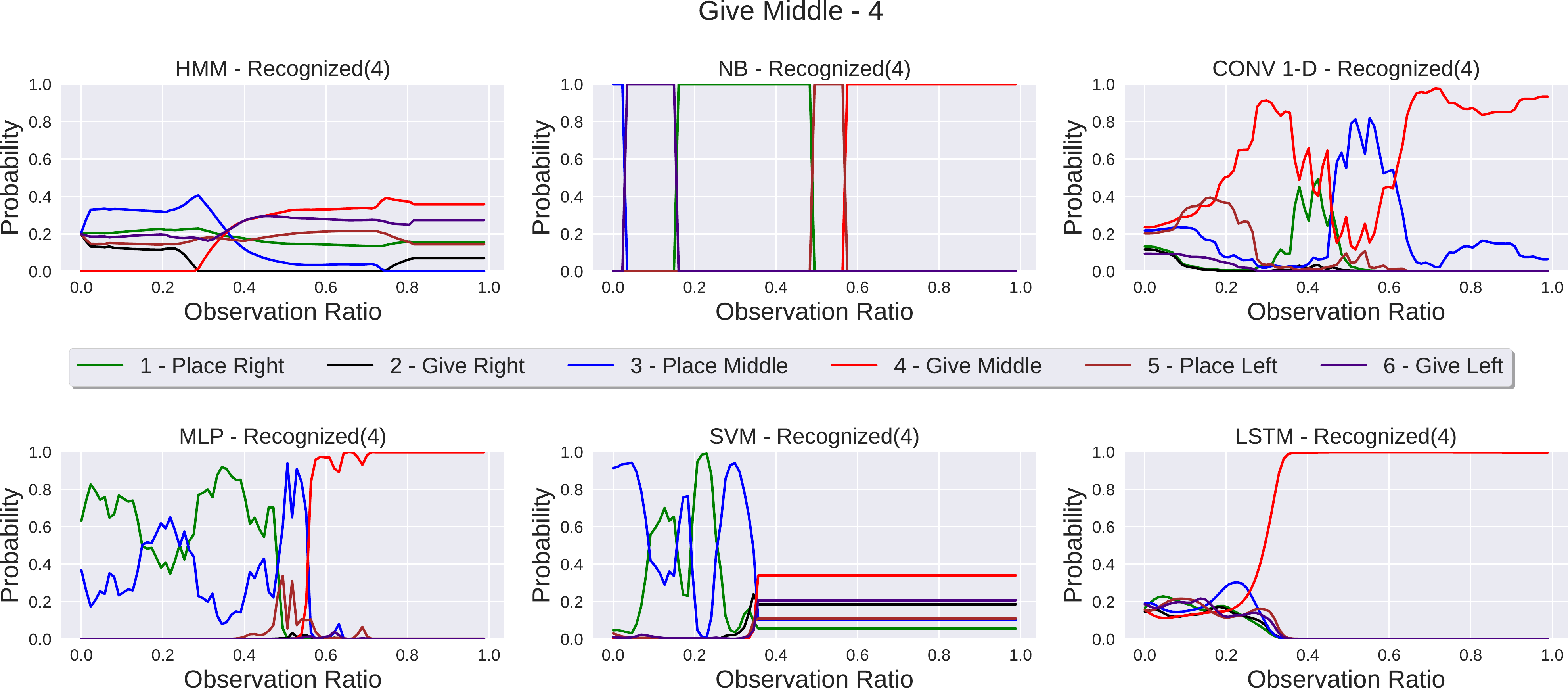}
\caption{Evolution of an action \textit{give middle} with its respective prediction for each baseline model and the proposed DLSTM. All models are trained on the original dataset (6 actions) with movement and head information.}
\label{fig:evolution_class}
\end{figure*}

As can be seen in Table~\ref{tab:acc_baselines}, the classical models, used as baselines, offer a satisfactory result when one analyzes only the accuracy obtained at each observation ratio. However, as mentioned before, this type of analyzis brings poor conclusions about the model's capacity to anticipate the actions. A good anticipation model should increase the differentiation between classes, while the number of observations also increases. Therefore, let's look at the graphs in Fig.~\ref{fig:evolution_class}, where we have the distribution of probabilities between classes for each observation ratio of an action \textit{give middle}. Note that the baseline models fail to accumulate knowledge about the performing actions (NB, HMM, MLP, CONV-1D, and SVM) or even respond over-confidently (NB) about prediction. In HMM the distribution is almost uniform, making it difficult to determine a proper probability to be used as a decision-making threshold. Naive Bayes responds with high confidence even at the beginning of the action. So it is not a reliable model to be used in anticipation tasks. The other three models, MLP, SVM, and CONV-1D are quite noisy and do not give confidence about the correct action. In contrast, the proposed DLSTM was capable of representing the evolution of the action, in such a way that clearly differentiates one action from the others while more observations are provided.

For a more complete analyzis, Table~\ref{tab:anticipation_acc} brings the anticipation accuracy of each model calculated by Eq.~\eqref{eq:acc_ant_act}. As we can see, the baseline models, even been quite effective in the recognition task, fail when applied to anticipation task. The max accuracy achieved by the HMM was $76.19\%$ when using a threshold of $0.38$. With such a small threshold, it is difficult to trust in this model, once at least one of the remaining classes can reach close values (e.g., $0.37$). The best among the other baseline models (MLP) achieved only $61.90\%$ with a threshold of $0.94$. While Naive Bayes could not even anticipate an action due to its over-confidence. On the other hand, DLSTM was capable of anticipating $95.25\%$ of actions when applying a threshold of $0.90$. All these pieces of evidence show us the superiority of the proposed model compared with the baselines.

\begin{table}[]
\tiny
\centering
\caption{Maximum anticipation accuracy (second column) obtained by each model (first column) when applying the probability threshold (last column). To achieve the corresponding accuracy the model needed on average an observation ratio such that specified in third column. As the NB model is extremely over-confident in its predictions, it is not suitable to anticipate  actions.}
\label{tab:anticipation_acc}
\begin{tabular}{lccc}

\hline
\multicolumn{1}{c}{Model} & \begin{tabular}[c]{@{}c@{}}Anticipation \\ Accuracy\end{tabular} & \begin{tabular}[c]{@{}c@{}}Mean Observation \\ Ratio Needed\end{tabular} & \begin{tabular}[c]{@{}c@{}}Probability\\  Threshold\end{tabular}  \\ \hline \hline
HMM  & 76.19\% &0.45 & 0.38 \\
NB   & \multicolumn{3}{c}{It is not possible to anticipate}\\
CONV-1D  & 47.62\% &0.69 & 0.91 \\
SVM  & 57.14\% &0.55 & 0.68   \\
MLP  & 61.90\% &0.60 & 0.94    \\
DLSTM  & 95.25\% &0.40 & 0.90  \\
\end{tabular}
\end{table}

Another import result is that our proposal, even using 2D skeleton joints extracted from images, outperforms \citep{schydlo2018anticipation}, that used eye gaze and 3D pose (Tab.~\ref{tab:acc_baselines}). With \textit{movement + head} information we achieved $90\%$ of accuracy with less than $40\%$ of observations. On the other hand, the authors' model, in \citep{schydlo2018anticipation}, achieves $90\%$ of accuracy after more than $50\%$ of observations. As discussed before, an effective anticipation model must also be an effective recognizer. Our model $ DLSTM_{6mh}$ recognized all actions at the last observation ($100\%$ of average accuracy); meanwhile, their model achieved a maximum of $97.5\%$ with an observation ratio $0.60$ and decreased to $87\%$ with an observation ratio $1.0$. Therefore, their model did not recognize all actions in the dataset.

In addition to the results above, $DLSTM_{6mh}$ can anticipate an action three frames before than $DLSTM_{6m}$, on average. As the video has a sample rate of 30Hz, this anticipation corresponds to 100ms, which is greater than the 92ms presented in \citep{schydlo2018anticipation} when comparing \textit{pose} with \textit{pose+gaze}. As such, besides outperforming \citep{schydlo2018anticipation}, our proposal was able to solve the action recognition problem in the Acticipate dataset and improve the action anticipation results.

\subsection{Deterministic models in the extended dataset}

\begin{figure} [h!]
\centering
\includegraphics[width =0.8\linewidth]{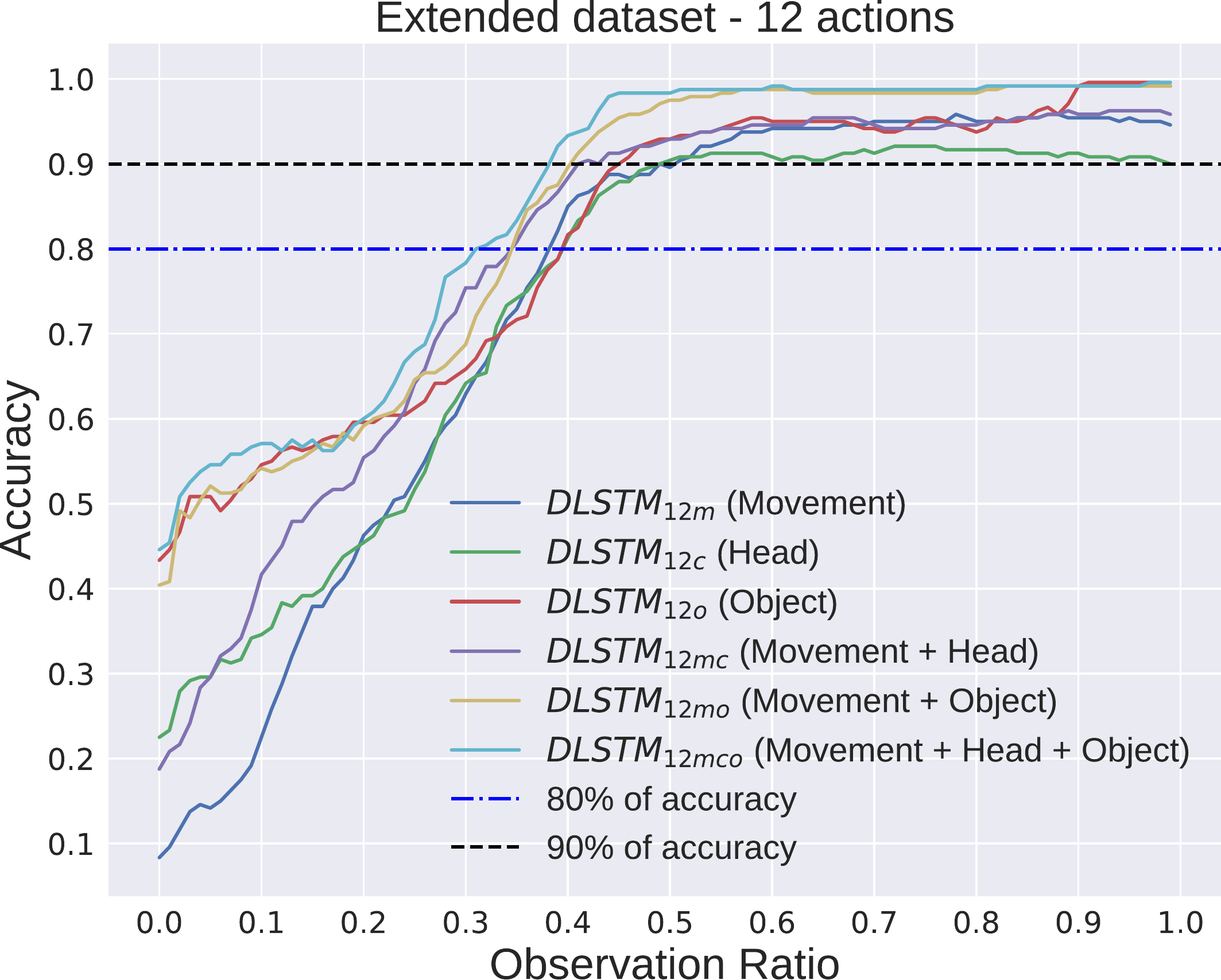}
\caption{Results for deterministic models in the extended dataset (12 actions). Each experiment is the same model trained with different input data.}
\label{fig:complete_12}
\end{figure}

Fig.~\ref{fig:complete_12} shows the results for the 6 experiments (from 8 to 13) using the extended version of the  dataset. We can observe that $DLSTM_{12m}$, $DLSTM_{12h}$, and $DLSTM_{12mh}$ did not achieve $100\%$ accuracy at the last observation. This result shows that they were unable to separate actions properly, even when using head information. The 6 new actions are the only difference between $DLSTM_{6mh}$ and these three models. As such, when the dataset was divided into more actions, more ambiguities were generated among them. Therefore, these results support our first hypothesis: more actions are likely to cause more ambiguities.

The models that use object information ($DLSTM_{12o}$, $DLSTM_{12mo}$, $DLSTM_{12mho}$) were able to recognize all actions (accuracy at 100\% of observations). Further, they achieved better results in the anticipation task. They start with more than $40\%$ of accuracy at the first observation, and the model with complete context ($DLSTM_{12mho}$) achieves $98\%$ of accuracy at the observation ratio of $0.42$. Therefore, actions with similar movements can be distinguished better when using context information. So, this result supports our second hypothesis: context information can help to distinguish different actions represented by similar movements. We can also see how this last model was able to extract relevant information from the object position and head points. Even though the object is represented by only two values (a 2-dimensional point), as we suppose, it provided an important information about the actions to the model. This result shows the efficacy of our feature embedding process.

\begin{figure} [h!]
\centering
\includegraphics[width =0.8\linewidth]{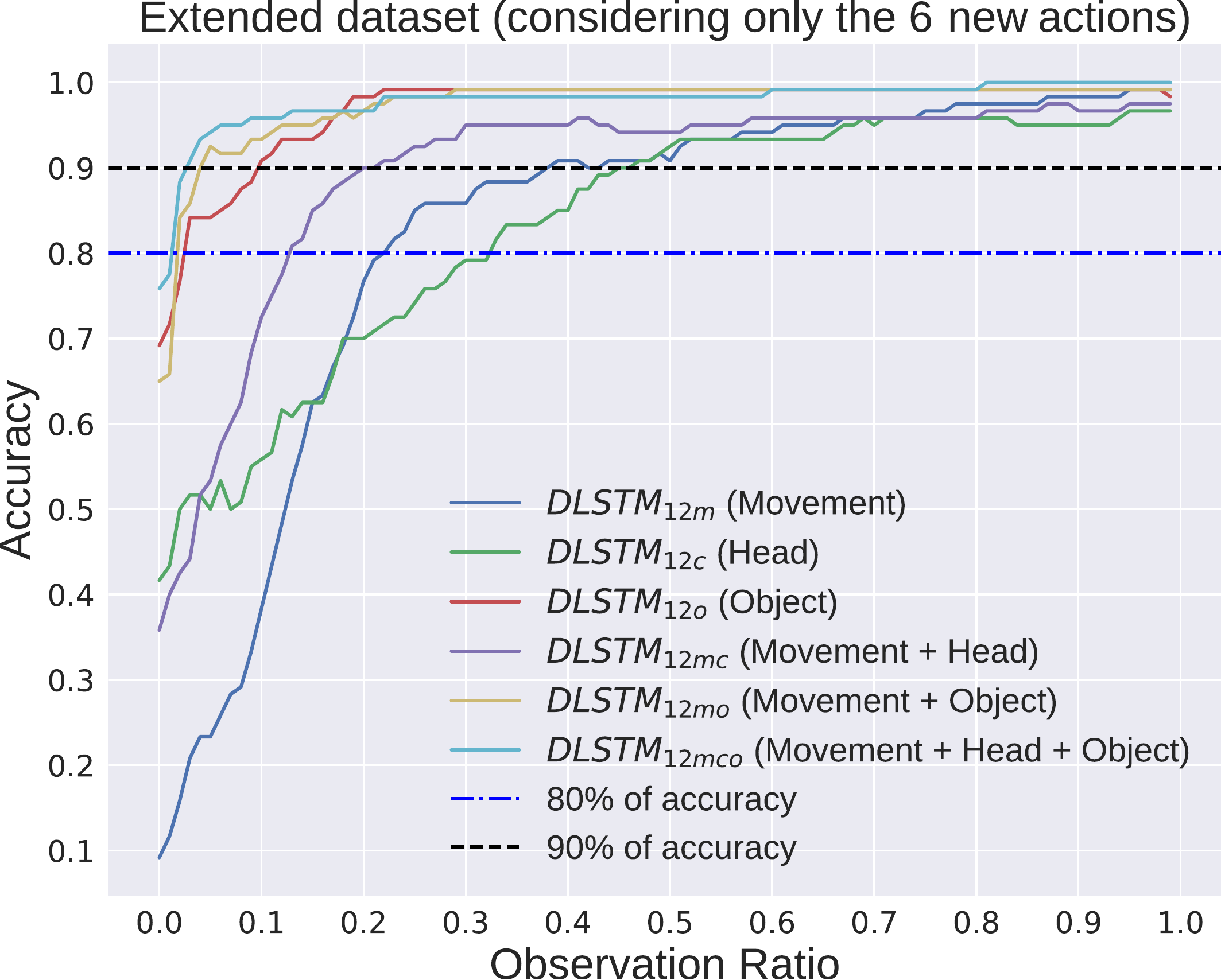}
\caption{Results for deterministic models in the extended dataset but cosidering only the 6 new actions.}
\label{fig:complete_last6}
\end{figure}

As mentioned above, after using the object information, the model might be able to anticipate some actions after a few observations. For better visualization, Fig.~\ref{fig:complete_last6} illustrates the accuracy of the same 6 models for the 6 new added actions (receive and pick (left, middle, right)). The models that use object context start with a classification accuracy greater than $65\%$ and achieve $90\%$ of accuracy after less than $10\%$ of observations. The best model reaches $95\%$ of accuracy with less than $5\%$ of observations, on average. In terms of frames, for the Acticipate dataset, that corresponds to an average of 4 frames. These results support our statement about the importance of object information for these 6 new actions. 

\begin{figure*} [h!]
\centering
\includegraphics[width =1.0\linewidth]{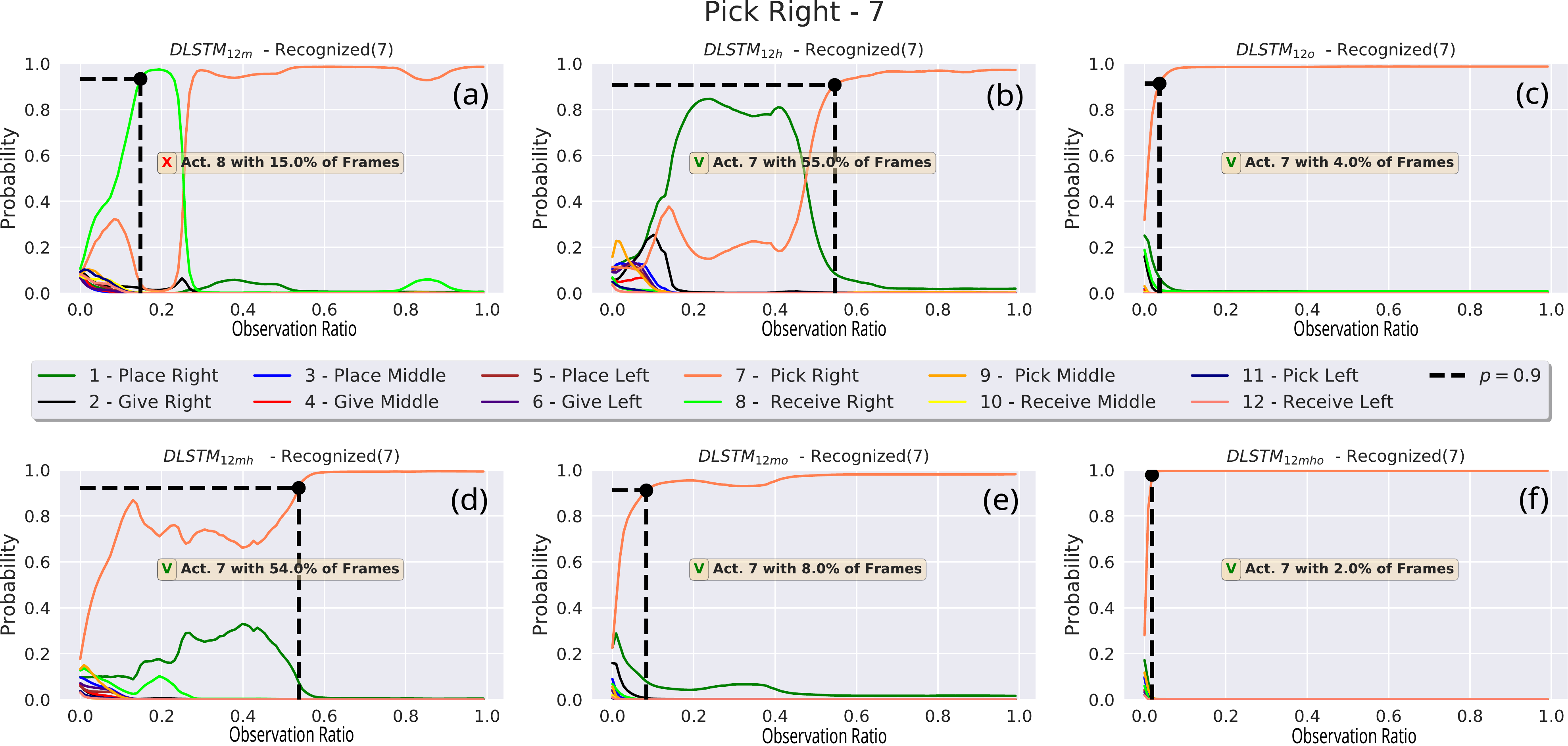}
\caption{Evolution of sample of a \textit{pick right} action for the 6 deterministic models.}
\label{fig:evolution_pick_rigkt}
\end{figure*}

To measure the anticipation accuracy, we use the Eq.~\ref{eq:acc_ant_act} with a threshold $p=0.9$ for the 6 models. Fig.~\ref{fig:evolution_pick_rigkt} presents the evolution of an action \textit{pick right} after passing throughout the 6 models. The charts illustrate how the model that uses only movement ($DLSTM_{12m}$) made a mistake in its anticipation. This mistake can be caused by the overconfidence of the model when anticipating ambiguous actions. Other models, which use part/complete context information, anticipated the action correctly. Notice that the model with full context (\textit{head + object}), anticipated the action after observing only $2\%$ of the data sequence (two frames in its corresponding video). Another interesting result is that the models confused those classes we supposed they would. After analyzing the videos, one can notice that action \textit{pick right} has similar movement to \textit{place right, give right} and \textit{receive right}, and similar gaze to \textit{place right}. Thus, $DLSM_{12m}$ \textit{pick right} mistook for \textit{receive right} and $DLSM_{12h}$ was not certain about \textit{pick right} and \textit{place right}. These characteristics appear in almost all predictions. 

\begin{figure} [h!]
\centering
\includegraphics[width =0.8\linewidth]{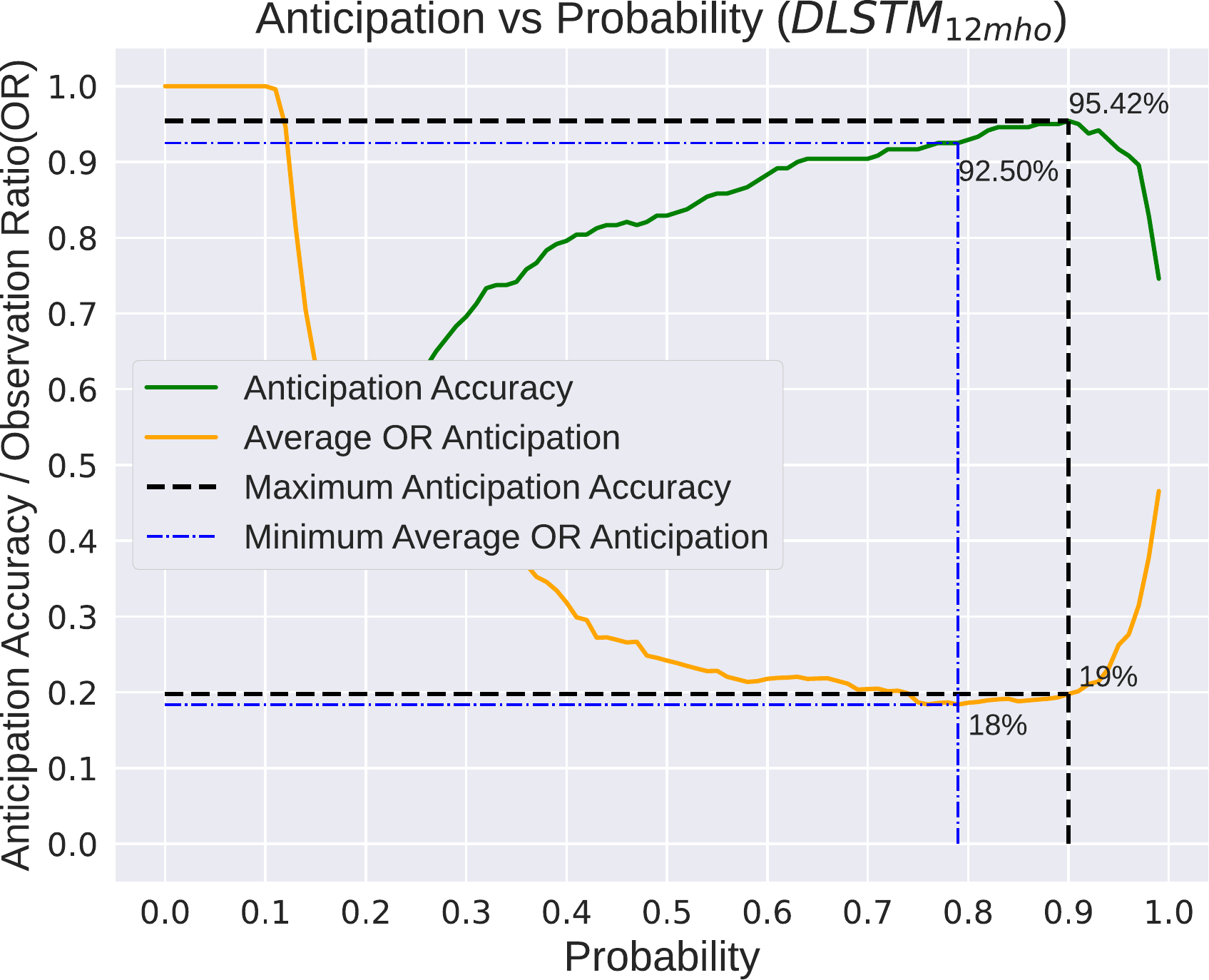}
\caption{Variation of anticipation accuracy and average observation ratio w.r.t probability threshold. }
\label{fig:probability_threshold}
\end{figure}

To highlight the trade-off between the threshold and the anticipation accuracy, the chart in Fig.~\ref{fig:probability_threshold} presents the variation of anticipation accuracy and the percentage of observations w.r.t threshold $(p)$. In the chart, we see that when $p=0.9$, $DLSTM_{12mho}$ can anticipate correctly $95.42\%$ of actions by using, on average, $19\%$ of the video sequence. Each action in Acticipate dataset has an everage of 79 images. Thus, this $19\%$ in observation ratio corresponds to an average of $15$ frames of a video. On the other hand, by aiming the minimum number of observations, with $p=0.8$, the model would anticipate correctly $92.50\%$ of actions by using, on average, $18\%$ of observation (14 frames).

\begin{figure} [h!]
\centering
\includegraphics[width =0.8\linewidth]{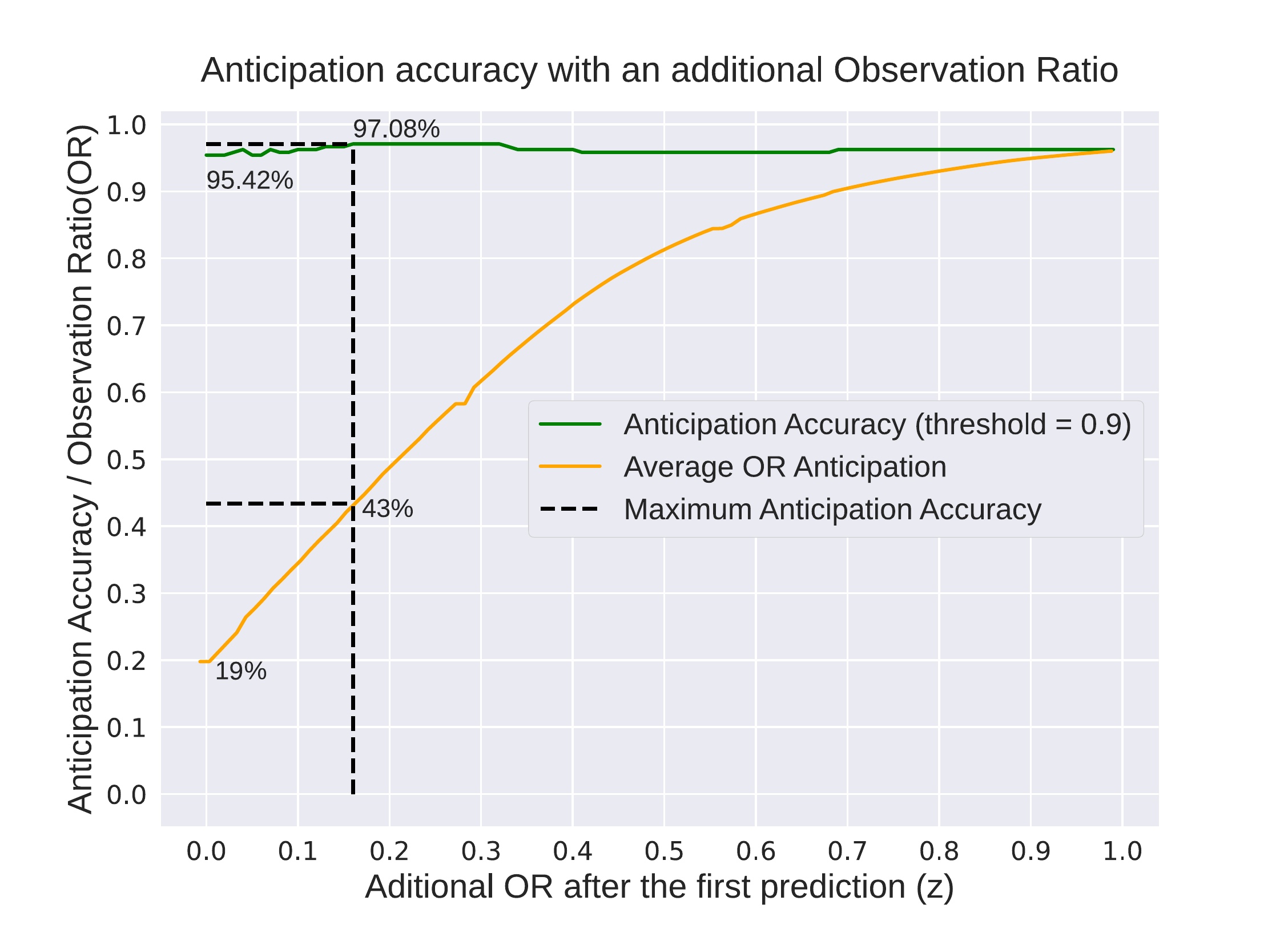}
\caption{Variation of anticipation accuracy and average observation ratio w.r.t additional observation ratio after anticipation. If in time $t$ the max probability exceeds the 0.9, the model must wait for more $z$ observations in order to conforms its prediction.}

\label{fig:duration_threshold}
\end{figure}

With $p=0.9$, function $g$ in Eq.~\eqref{eq:anticipation_threshold} is not able to consider overconfidence in the model prediction, which may generate many false-positives. As discussed in Sec.~\ref{sec:intuitions}, a possible solution to reduce the number of false-positives is to force the model to wait for more $z$ observations to reaffirm its prediction. The problem with this approach is that $z$ is a new parameter that can harm the anticipation, and must be chosen carefully. Fig.~\ref{fig:duration_threshold} illustrates how anticipation accuracy ($p = 0.9$) and average observation ratio vary w.r.t $z$, where $z$ is the additional observation ratio after anticipation. The best anticipation accuracy ($97.08\%$) is achieved when $z=0.18$. In other words, the model needs to wait on average for more $18\%$ of observations to achieve an anticipation accuracy of $97.08\%$. Comparing with previous results, the gain of less than 2\% in the accuracy cost an increase of more than the double of observations to the anticipation time (passing from $19\%$ to $43\%$). Furthermore, the minimum observation ratio necessary to anticipate any action is now $18\%$, even for less ambiguous actions, such as those presented previously in Fig.~\ref{fig:sample_ambuiguities_object} and Fig.~\ref{fig:evolution_pick_rigkt}. As conclusion, besides the fact that the choice of $z$ inserts a new trade-off in the project (accuracy \textit{vs} observation ratio), it does not provide an effective way to improve the action anticipation task.
 
\subsection{Stochastic models} 
The results of the Bayesian models $LSTM_{MC}$, $BLSTM_{VD}$ and $BLSTM_{BBB}$ will be compared to $DLSTM_{12mho}$, our best deterministic model for the extended dataset. During prediction time, we fed each Bayesian model 20 times with the same observation $\xx_t$, which corresponds to an MC sampling with $S = 20$. Next, by applying Eq.~\eqref{eq:mutual_information} over the $S$ predictions, we measured the epistemic uncertainty of each model prediction, concerning the observation $\xx_t$. Then, we could use Eq.~\eqref{eq:uncert_ant_func} to anticipate the action or not. 

\begin{figure} [h!]
\centering
\includegraphics[width =0.8\linewidth]{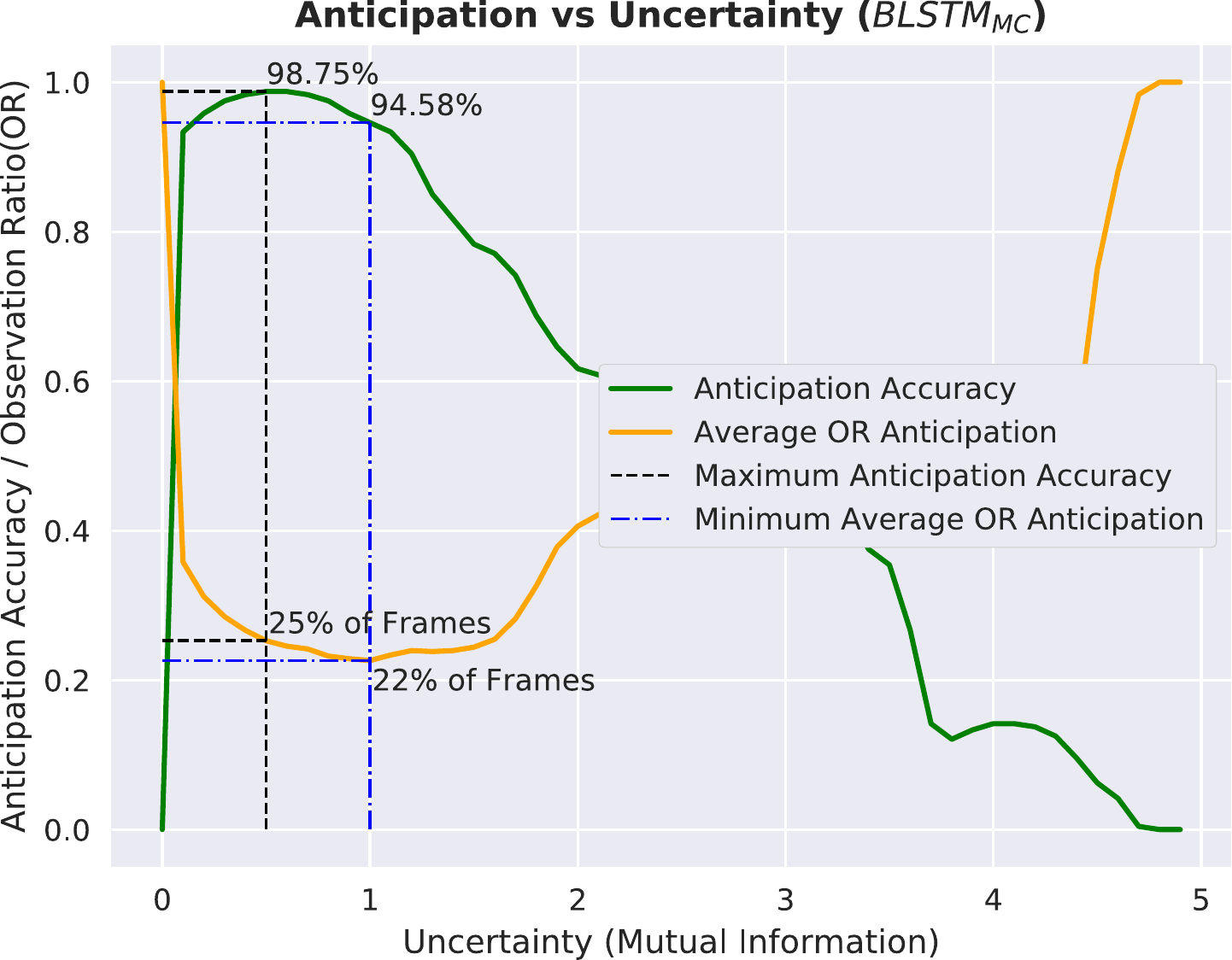}
\caption{Variation of anticipation accuracy and average observation ratio w.r.t uncertainty threshold for MC dropout model }
\label{fig:uncertainty_mc_threshold}
\end{figure}

The Bayesian models also recognized all actions in the extended dataset. Furthermore, they achieved better results in anticipation accuracy than $DLSTM_{12mho}$, even if it waits for an observation ratio of $z=0.18$. By applying the same procedure in Fig.~\ref{fig:probability_threshold}, we could choose a threshold value to be used in each model. Therefore, for each model, the anticipation threshold was chosen by analyzing the variation of anticipation accuracy and the average observation time w.r.t the uncertainty value. Fig.~\ref{fig:uncertainty_mc_threshold} shows this comparison for $BLSTM_{MC}$. 

Table \ref{tab:models_results} compares the results of the Bayesian models with our best deterministic model ($DLSTM_{12mho}$). Note that $BLSTM_{MC}$ achieves the best anticipation accuracy ($98.75\%$) using the uncertainty threshold $u=0.5$. However, $BLSTM_{VD}$ and $BLSTM_{BBB}$ also achieves satisfactory results: with $u=0.5$, $BLSTM_{VD}$ achieved $98.33\%$ of anticipation accuracy, and with $u=0.3$ $BLSTM_{BBB}$ achieves $97.08\%$.

 \begin{table*}[h!]
 \centering
 \small
 \caption{Results of stochastic models and the best deterministic model.}
\label{tab:models_results}

\begin{tabular}{lccc}
\hline
\multicolumn{1}{c}{Model} & Parameter      & \begin{tabular}[c]{@{}c@{}}Anticipation \\ Accuracy\end{tabular} & \begin{tabular}[c]{@{}c@{}}Average \\ Observation Ratio\end{tabular} \\ \hline 
$DLSTM_{12mho}$      & $p = 0.9$ / $z = 0.0$          & $95.42\%$                             &$19\%$                               \\
$DLSTM_{12mho}$      & $p = 0.79$ / $z = 0.0$          & $92.50\%$                             &$\textbf{18\%}$                              \\
$DLSTM_{12mho}$      & $p = 0.9$ / $z = 0.18$         & $97.08\%$                             &$43\%$                               \\
$BLSTM_{MC}$       & $u = 0.5$                        & $\textbf{ 98.75}\%$                   &$25\%$                               \\
$BLSTM_{MC}$       & $u = 1.5$                        & $94.58\%$                             &$22\%$                               \\
$BLSTM_{VD}$       & $u = 0.5$                        & $98.33\%$                             &$26\%$                               \\
$BLSTM_{VD}$       & $u = 1.5$                        & $85.42\%$                             &$20\%$                               \\
$BLSTM_{BBB}$       & $u = 0.3$                       & $97.08\%$                             &$25\%$                              \\
$BLSTM_{BBB}$       & $u = 1.3$                       & $93.33\%$                             &$20\%$                              \\\hline                    
\end{tabular}
\end{table*}

Considering the minimum number of observations necessary for a good anticipation accuracy, $DLSTM_{12mho}$ gives the best result. On average, with $p=0.79$, it needs to receive $18\%$ of observations to achieve an anticipation accuracy of $92.50\%$. However, even needing less observations ($18\%$) with $p=0.79$, it presents less accuracy ($92.50\%$) than considering more observations at a threshold of $p=0.9$, which achieves an accuracy of $95.42\%$. Therefore, we can see that there is a tradeoff of accuracy for a less number of observations. In summary, the best model achieved $98.75\%$ after observing, on average, $25\%$ of the action ($BLSTM_{MC}$). An increase of $6.67\%$ in the accuracy with a cost of only $7\%$ in extra observations. Much better than using $z$ in the deterministic model, where an increase of less than $2\%$ costs $24\%$ on extra observations. Besides, we do not need to choose more than one hyperparameter, only the uncertainty threshold $u$.

\subsection{Discussions}

As we mentioned in the previous sections, the model must have a short anticipation time for human-machine interaction and be accurate in its prediction. For $BLSTM_{MC}$, it needs $25\%$ of observations to achieve its best prediction value, which indeed is not a high value. For instance, in a system based on images sampled at $30$Hz (ordinary cameras), an action that lasts $2$ seconds would be anticipated by such a model after elapsed on average $0.5s$ from its first frame. In other words, it might anticipate an action after the system observes, on average, $15$ frames. Therefore, once the model can be considered accurate in its prediction, the system has about $1.5s$ to make a right decision. 

As expected, our Bayesian models provided better results than deterministic ones with a small cost in additional observations. The overconfidence in model prediction decreases when waiting for more observations. However, as we could see, for deterministic models, this is a new parameter to be chosen ($z$) and did not provide satisfactory results. On the other hand, by using uncertainty as a threshold, we have only one parameter to be chosen, and the model can achieve better results of accuracy with a small cost in the observation ratio. These results support our last hypothesis that: uncertainty is a more reliable and effective threshold to anticipate actions than probability values.

In our opinion, the MC dropout~\citep{gal2016dropout,gal2016rnn} and variational dropout~\citep{kingma2015variational_dropout} were the best models implemented in this work. Once dropout and local reparametrization can provide a different sample for each observation, a mini-batch with $S$ observations correspond to an MC sampling of size $S$, which helps the model posterior distribution inference. Besides, for prediction, we only need to create a mini-batch of size $S$, repeating the same observation, that favors parallel prediction in GPUs. On the other hand, the reparametrization trick does not take advantage of the mini-batch to make samples. Every observation in the mini-batch uses the same sampled weight.

As a consequence, in our experience, BBB models train slower than Bayesian dropout approaches, and, during prediction, it needs to run the model $S$ times with the same observation, which makes impractical to parallelize the prediction in GPUs. However, it seems that a significant advantage of BBB is the possibility of pruning the model by analyzing each parameter. As they are Gaussian distributions, the relation mean-variance can indicate if a parameter is required or maybe discarded \citep{graves2011practical,blundell2015weight}.

Finally, we could see that the proposed model outperformed the baselines, including \cite{schydlo2018anticipation}, even using less accurate information (2D vs. 3D pose and head joints vs. eye gaze). The results supported the raised hypotheses and showed how the uncertainty provided by Bayesian models is vital for action anticipation. Our proposal can be used in other datasets even though the presented results were acquired in a small collaborative dataset. In this sense, it is necessary to analyze the possible sources of context for each class and adapt our embedding layer to represent all the context data.

\section{Conclusions and Future Works}
\label{sec:conclusions}
Machines need the capacity of anticipating actions to achieve effective interaction with humans. As such, the problem of action anticipation is drawing substantial research attention in recent years. Although many works have explored the issue, they do not provide a concise explanation about the importance of context in anticipating actions. They do not discuss how to handle the problem of the uncertainty inherent in this kind of task and how to make decisions in a real-time situation.

We propose a decision-making criterion based on the uncertainty provided by a stochastic (Bayesian) LSTM model that can practically be used for action anticipation tasks. By selecting the action that minimizes the uncertainty, our model improves the action anticipation performance compared with the conventional class-likelihood maximization (i.e., deterministic model).

Considering arm motion as the primary source of information for action anticipation, we evaluate the influence of two additional (contextual) sources of information in the Acticipate dataset: gaze and object attributes.
When considering all information sources in our stochastic LSTM, we achieved $100\%$ of average accuracy in the action recognition task and $98.78\%$ of average accuracy in the action anticipation task, outperforming previous results. Thus, our model serves both action recognition and anticipation purposes, while needing only $25\%$ of the observations, on average, to anticipate each action. The results also show the evident importance of context for the anticipation task, since the actions that depend on the eye gaze information or the object position had impressive improvement in their anticipation time contrasted with \citep{schydlo2018anticipation}. For instance, actions that depend exclusively on the object information are anticipated precociously, some of them with only two observations.

Our work extends the current state-of-the-art and results in action anticipation, for small collaborative datasets. Also, our proposal uses context information to improve the classification probability, and the uncertainty as the decision-making criterion that can be used with any other probabilistic model.

As future work, we aim to increase the collaborative setup complexity by adding more objects to each action and designing a collaborative scenario where the performed actions depend on more than one object. Another important issue to be addressed is how to extract proper context from general datasets. In such a way, we could use this proposal to solve more complex anticipation tasks.

\section*{Acknowledgments}
This study was financed in part by the Coordena\c{c}\~{a}o de Aperfei\c{c}oamento de Pessoal de N\'{i}vel Superior - Brasil (CAPES), PDSE/Process nº 88881.188840/2018-01, Finance Code 001. The authors also would like to acknowledge the support from NVIDIA Corporation through the donation of the Titan V GPU used in this research.

\section*{Bibliography}
\bibliography{referencies}

\end{document}